\documentclass[a4paper,fleqn]{cas-sc}
\usepackage{fancyhdr} 
\usepackage{caption}
\usepackage[authoryear,longnamesfirst]{natbib}
\usepackage{subcaption}
\usepackage{tikz}
\usetikzlibrary{shapes.geometric, arrows}
\usepackage{pdflscape}
\usepackage{amsmath}
\usepackage{float}
\usepackage{graphicx}
\def\tsc#1{\csdef{#1}{\textsc{\lowercase{#1}}\xspace}}
\tsc{WGM}
\tsc{QE}
\tsc{EP}
\tsc{PMS}
\tsc{BEC}
\tsc{DE}

\begin{document}
\let\WriteBookmarks\relax
\let\printFirstPageNotes\relax
\def\floatpagepagefraction{1}
\def\textpagefraction{.001}

\title [mode = title]{Deep Learning with Self-Attention and Enhanced Preprocessing for Precise Diagnosis of Acute Lymphoblastic Leukemia from Bone Marrow Smears in Hemato-Oncology}

\author[inst1]{\small Md. Maruf}
\author[inst1]{\small Md. Mahbubul Haque}
\author[inst1]{\small Bishowjit Paul}

\affiliation[inst1]{organization={Department of Mechatronics Engineering},
            addressline={Rajshahi University of Engineering \& Technology},
            city={Rajshahi},
            country={Bangladesh}}

\begin{abstract}
Acute Lymphoblastic Leukemia (ALL) is a prevalent hematological malignancy that affects both the pediatric and adult populations. Early and accurate detection, along with precise subtyping, is critical to improve patient outcomes and inform optimal therapeutic strategies. However, conventional diagnostic approaches are often complex, time-consuming and subject to human error, leading to delays in timely intervention. This study introduces an advanced deep learning-based framework for the automated diagnosis of ALL from bone marrow smear images, eliminating the dependence on manual interpretation. The proposed methodology incorporates a robust image preprocessing pipeline that enhances the quality of bone marrow smear images for optimized input into convolutional neural networks (CNNs), thereby significantly improving diagnostic accuracy and computational efficiency.As a key innovation, the framework integrates a Multi-Head Self-Attention (MHSA) mechanism within the CNN architecture to capture long-range dependencies and contextual relationships between cell features, enabling a more nuanced representation of complex morphological patterns in leukemic cells. Furthermore, Focal Loss is used during training to address the challenge of class imbalance, ensuring that minority class samples are effectively learned and contributing to the improvement of model generalization. Among the architectures evaluated, the enhanced VGG19 model, equipped with the MHSA mechanism and trained with Focal Loss, achieved a remarkable accuracy of 99. 25\%, exceeding the previously established benchmark of 98.62\% set by ResNet101. This superior performance underscores the efficacy of the proposed innovations in model architecture, training strategy, and image processing.By delivering a highly accurate, interpretable, and clinically viable diagnostic model, this research makes a significant contribution to the domain of automated ALL detection. The integration of self-attention and loss optimization techniques not only advances the state of the art in bone marrow smear analysis but also holds substantial promise for accelerating and refining leukemia diagnostics in real-world clinical settings.
\end{abstract}


\maketitle

\section{Introduction}

Acute Lymphoblastic Leukemia (ALL) is a prevalent and potentially fatal hematologic malignancy that primarily affects blood cells and the bone marrow. Recent years have seen a concerning increase in both the number of cases and mortality rates, impacting millions of children and adults worldwide. Although ALL predominantly affects children and adolescents \cite{vadillo2018t}, adult cases, though rarer, are also observed. Early and accurate detection is crucial for improving survival rates and guiding treatment strategies. Traditionally, hematologists diagnose ALL through manual examination of bone marrow smear images, a process that is labor intensive and prone to human error, leading to delays and potential misdiagnoses \cite{wu2020hematologist}. In contrast, computer-aided diagnostic systems, particularly those based on deep neural networks, offer the potential to revolutionize early detection and diagnostic accuracy \cite{rezayi2021timely}.

ALL typically originates from genetic mutations in the precursor cells of lymphocytes within the bone marrow, affecting normal cell growth and causing the uncontrolled proliferation of abnormal lymphocytes or lymphoblasts \cite{rosmarin2019leukemia}. As these leukemic cells multiply, they suppress the production of healthy blood cells and can spread to other organs such as the kidneys, liver, brain, and heart, leading to additional complications and symptoms such as fatigue, frequent infections, and easy bruising \cite{sampathila2022customized, jiwani2023pattern}.

\begin{figure}[h] \centering \begin{minipage}{0.49\textwidth} \centering \includegraphics[width=\linewidth]{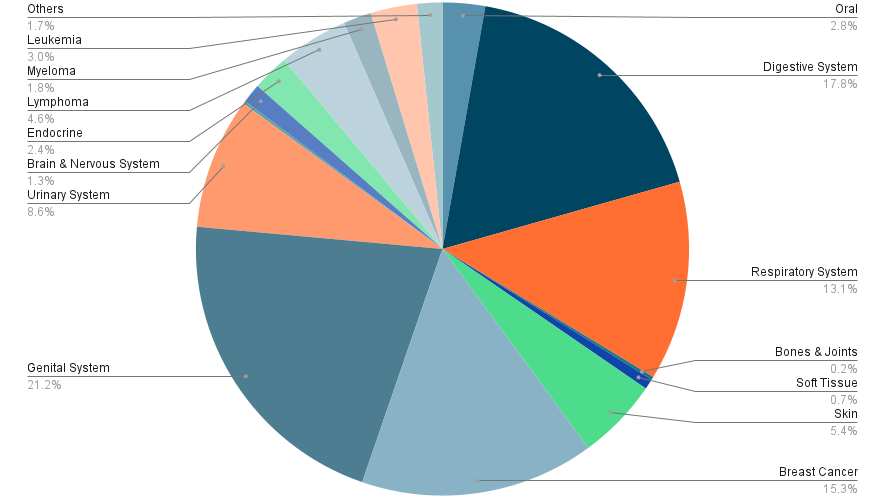} \label{fig:diagnosed_cases} \end{minipage}\hfill \begin{minipage}{0.49\textwidth} \centering \includegraphics[width=\linewidth]{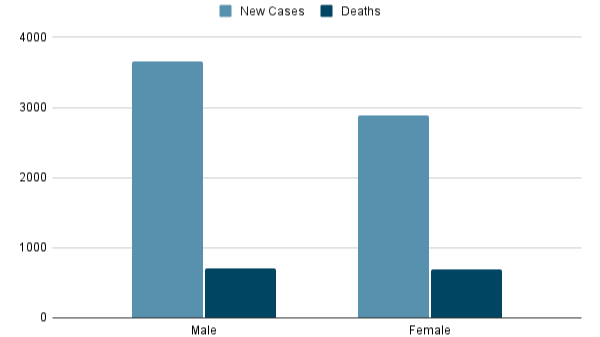} \label{fig:ALL_cases_deaths} \end{minipage} \caption{(a) Percentage of diagnosed cancer cases. (b) Estimated ALL new cases and deaths in the USA, 2023.} \label{fig:combined} \end{figure}

ALL is the most common cancer in children, particularly those between the ages of 2 and 5, accounting for approximately 27\% of all pediatric malignancies \cite{onyije2022environmental}. The overall survival rate for pediatric ALL patients is around 90\% \cite{lejman2022genetic}, while in adults, ALL is a rare condition, comprising only 1\% of adult malignancies, with a significantly lower survival rate of approximately 20\% \cite{juluri2022asparaginase}. As depicted in Figure 1, the incidence and survival rates underscore the stark contrast between pediatric and adult ALL cases. According to the American Cancer Society, in 2023, there will be an estimated 6,540 new cases of ALL in the United States, with 1,390 related deaths \cite{siegel2023cancer}. The Surveillance, Epidemiology, and End Results (SEER) program reports a relative survival rate of 71.3\% for ALL patients diagnosed between 2013 and 2019. Furthermore, the International Agency for Research on Cancer (IARC) of the WHO recorded 437,033 leukemia cases and 303,006 deaths globally in 2022 \cite{sheykhhasan2022use}. These statistics highlight the global burden of ALL and its profound impact across various age groups.

ALL is an aggressive cancer that can progress rapidly if not detected early \cite{chand2022novel}. Timely and accurate diagnosis is essential to improving treatment outcomes. However, current diagnostic methods—ranging from clinical assessments to advanced imaging and molecular testing—are often time-consuming, subjective, and prone to errors \cite{horvath2022peripheral}. Deep learning technologies present a transformative solution by enabling automated and precise detection of ALL, reducing the reliance on manual interpretation. This research aims to develop a deep learning-based system designed to enhance the accuracy of ALL detection from peripheral blood smear images, complementing the work of hematopathologists rather than replacing them.

The primary objectives of this study are to preprocess peripheral blood smear images to highlight relevant features, develop a deep learning model capable of classifying ALL at various stages, minimize classification errors, and evaluate the system's performance against existing diagnostic methods. The key contributions of this research include the design of a novel deep learning framework for ALL detection from blood smears, addressing challenges such as data variability and limited sample availability. The model's performance is enhanced by integrating custom layers into a pre-trained neural network, improving its adaptability and minimizing overfitting. Furthermore, the system places a strong emphasis on interpretability, ensuring its clinical applicability while adhering to ethical standards, including patient care and data privacy.

This paper explores the use of deep learning for the detection of acute lymphoblastic leukemia (ALL). It introduces the context of ALL and the rationale for the study, reviews existing research while identifying gaps, and discusses the dataset utilized. The methodology section outlines the preprocessing steps for peripheral blood smear images and the application of a hybrid deep learning model for classification. The results are then analyzed and compared with previous methods, followed by a discussion of findings and directions for future research.

\section{Acute Lymphoblastic Leukemia}
The distinct collection of proteins on the surface of leukemia cells, known as immunophenotypes, is crucial for their identification. According to the World Health Organization (WHO), acute lymphoblastic leukemia (ALL) is categorized into two primary subtypes based on these immunophenotypic profiles: B-cell and T-cell lymphoblastic leukemia \cite{anilkumar2022automated}.
\begin{figure}[h]
	\centering
	\includegraphics[width=\textwidth]{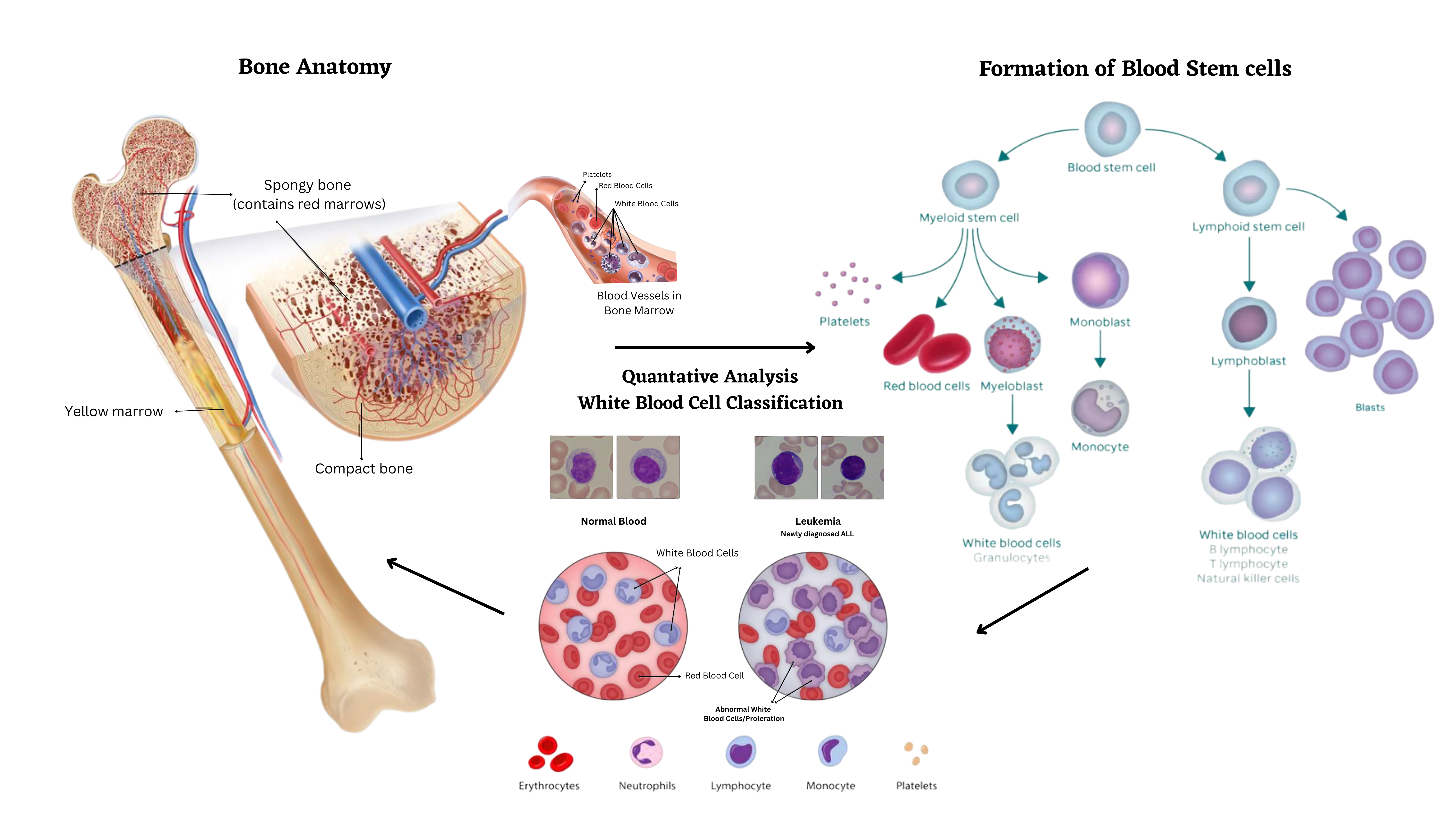}
	\caption{ Formation of B-lymphocytes and T-lymphocytes from blood stem cells}
\end{figure}

Hematopoietic stem cells in the bone marrow are the progenitors of both B-lymphocytes and T-lymphocytes \cite{grvcevic2023interactions}. The differentiation of B-lymphocytes involves the expression of specific transcription factors and cell surface markers that help in distinguishing between various developmental stages \cite{medina2022flt3}. Mutations occurring in DNA-synthesizing cells are implicated in the development of B-lymphocyte leukemia, which is the more common subtype of ALL.

T-lymphocytes originate from initial lymphoid progenitors that migrate from the bone marrow to the thymus, where they mature \cite{sam2021t}. During this maturation process, autoimmune T-cells are removed, and functional T-cells undergo positive selection \cite{teachey2019comparative}. The T-cell subtype of ALL is associated with mutations in the NOTCH1 gene, which plays a critical role in T-cell development and is linked to the less common but more severe T-cell leukemia. This subtype is more prevalent in adults than in children, representing roughly 25\% of ALL cases in adults. The immunophenotypic differentiation of ALL subtypes not only aids in accurate diagnosis but also informs treatment strategies and prognostic assessments, emphasizing the importance of understanding these cellular and molecular characteristics in the management of leukemia.

\section{Computer-Aided Diagnosis}
Computer-Aided Diagnosis (CAD)  is a transformative technology in healthcare that enhances diagnostic accuracy and efficiency \cite{baaa010}. In the detection of Acute Lymphoblastic Leukemia (ALL), CAD systems assist hematologists and pathologists by identifying subtle cellular anomalies in medical imaging data, such as bone marrow samples \cite{chan2020deep}. Utilizing advanced algorithms and machine learning, CAD systems improve diagnostic precision and provide additional insights without replacing human expertise \cite{ahmad2021artificial}.

A significant advantage of CAD is its ability to detect ALL early, which is crucial for timely intervention and better patient outcomes. These systems offer consistent diagnostic assessments, reducing human error related to fatigue or varying conditions \cite{yanase2019systematic}. By processing large volumes of images efficiently, CAD systems streamline workflows, allowing medical professionals to focus on more complex cases and research \cite{najjar2023redefining}. Overall, CAD enhances early diagnosis, consistency, and patient care in the field of ALL detection.

\section{Literature Review}
Recent advancements in the automated classification of Acute Lymphoblastic Leukemia (ALL) cells from bone marrow smear images have predominantly leveraged deep learning techniques, particularly Convolutional Neural Networks (CNNs) and transfer learning models, due to their superior performance in feature extraction and classification tasks \cite{morid2021scoping}.

Atteia et al. \cite{atteia2022bo} proposed a Bayesian-optimized CNN framework that significantly improved classification accuracy through meticulous hyperparameter tuning. Anilkumar et al. \cite{anilkumar2022automated} demonstrated a model employing AlexNet and LeukNet, achieving 94.12\% accuracy in distinguishing between B-cell and T-cell leukemia, highlighting the effectiveness of customized architectures in enhancing detection performance.
\begin{table}[ht]
\centering
\label{tab:summary_of_research}
\caption{Summary of Previous Research on ALL Detection}
\begin{tabular}{|p{2cm}|p{3cm}|p{3cm}|p{3cm}|p{3cm}|}
\hline
\textbf{Reference} & \textbf{Proposed} & \textbf{Results} & \textbf{Limitation} & \textbf{Findings} \\
\hline
Atteia et al. \cite{atteia2022bo} & Optimized Bayesian CNN & Enhanced classification through hyperparameter tuning & None specified & Improved accuracy through Bayesian optimization \\
\hline
Anilkumar et al. \cite{anilkumar2022automated} & AlexNet and LeukNet for B-cell and T-cell classification & Achieved 94.12\% classification accuracy & No image segmentation or feature extraction used & Accurate classification without complex processing \\
\hline
Jha et al. \cite{jha2019mutual} & Hybrid model with Mutual Information and CNN & Achieved 98.7\% accuracy & Possible noise and data imbalance issues & High accuracy with hybrid approach \\
\hline
Sampathila et al. \cite{sampathila2022customized} & ALL-NET deep learning classifier & Achieved 95\% accuracy & Loss of information with excessive max pooling & High accuracy but limited in noisy images \\
\hline
Saeed et al. \cite{saeed2022deep} & Multi-Attention EfficientNet models with transfer learning & Enhanced model generalization with attention mechanisms & Small dataset size may limit robustness & Improved accuracy with advanced architecture \\
\hline
Vogado et al. \cite{vogado2018leukemia} & CNNs with SVM for classification & Effective without segmentation; used combined databases & Need for extensive image databases & Effective classification with a need for more data \\
\hline
Zhou et al. \cite{zhou2021development} & Cell counting and classification based on FAB criteria & Practical workflow with cell counting & Limited to FAB criteria, may need broader applications & Strong clinical relevance and cell counting functionality \\
\hline
Kasani et al. \cite{kasani2020aggregated} & Aggregation-based deep learning with NASNetLarge and VGG19 & Achieved 96.58\% accuracy & Data augmentation issues; resizing may lose information & High accuracy and fast decision-making \\
\hline
Jiang et al. \cite{jiang2021method} & ViT-CNN ensemble with Difference Enhancement Random Sampling (DERS) & Addressed class imbalance and noise & Small dataset size; noise influence & Promising accuracy, but needs dataset expansion \\
\hline
\end{tabular}

\end{table}

Jha et al. \cite{jha2019mutual} developed a hybrid model integrating Mutual Information for segmentation and a CNN classifier, attaining a remarkable accuracy of 98.7\%. Sampathila et al. \cite{sampathila2022customized} introduced the ALL-NET classifier, which achieved 95\% accuracy; however, the model exhibited limitations related to noise interference and feature loss, suggesting potential avenues for improvement.

Saeed et al. \cite{saeed2022deep} employed Multi-Attention EfficientNet models with transfer learning to enhance model generalization, showcasing the adaptability of transfer learning in various clinical datasets. Meanwhile, Vogado et al. \cite{vogado2018leukemia} utilized CNNs combined with Support Vector Machines (SVM) for classification without segmentation, emphasizing the critical need for extensive image databases to ensure robustness and reliability.

Zhou et al. \cite{zhou2021development} developed a methodology that mimics hematologist workflows by focusing on cell counting, promoting practical clinical applications of automated ALL detection systems. Additionally, Kasani et al. \cite{kasani2020aggregated} introduced an aggregation-based deep learning approach, achieving 96.58\% accuracy with enhanced decision-making efficiency and data augmentation techniques.

Jiang et al. \cite{jiang2021method} proposed a ViT-CNN ensemble model integrated with a novel sampling technique to effectively address challenges related to imbalanced and noisy datasets, underscoring the importance of robust model design for real-world applications.

By synthesizing these contributions, this research positions itself within the broader academic discourse, aiming to advance the field of automated ALL detection through the development of high-performance deep learning models.

\section{Dataset Description}
Diagnosing Acute Lymphoblastic Leukemia (ALL) is both costly and labor-intensive. Peripheral blood smear (PBS) images are essential for initial screening but pose challenges due to the non-specific symptoms of ALL, leading to potential diagnostic errors. Public datasets like ALL-IDB, C-NMC 2019, and the TCGA leukemia dataset are commonly used for ALL diagnosis and are available from cancer imaging repositories. Private datasets, created from clinical screenings, are smaller and require more effort to manage. Public datasets can be updated by authorized personnel, whereas private datasets are managed by their creators.

The Taleghani Hospital dataset includes 3242 PBS images from 89 suspected ALL patients \cite{aria2021acute}. After image processing and data augmentation, the dataset was divided into 6480 training images and 720 validation images. For testing purpose we separately created a totally unseen test dataset where each class has 100 samples. Features from the training images are used to train a pretrained model with a task-specific classifier, while features from the testing images are used for predictions. Model performance is evaluated using metrics like accuracy, precision, recall, and F1-score. Figure 3 (a) shows the sample distribution.
\begin{figure}[h]
    \centering
    \begin{minipage}{0.5\textwidth}
        \centering
        \includegraphics[width=\textwidth]{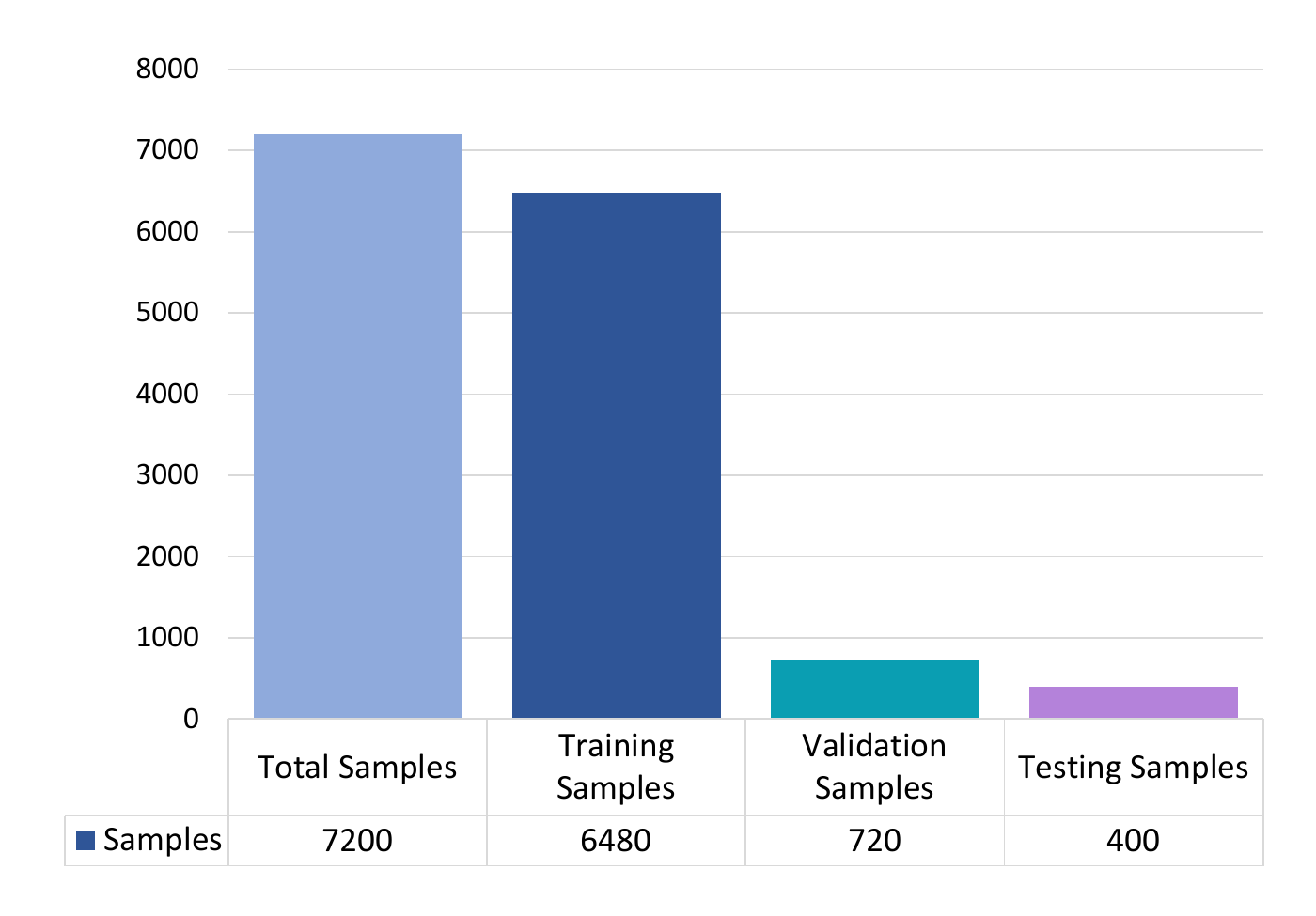}
        
    \end{minipage}%
    \begin{minipage}{0.5\textwidth}
        \centering
        \includegraphics[width=\textwidth]{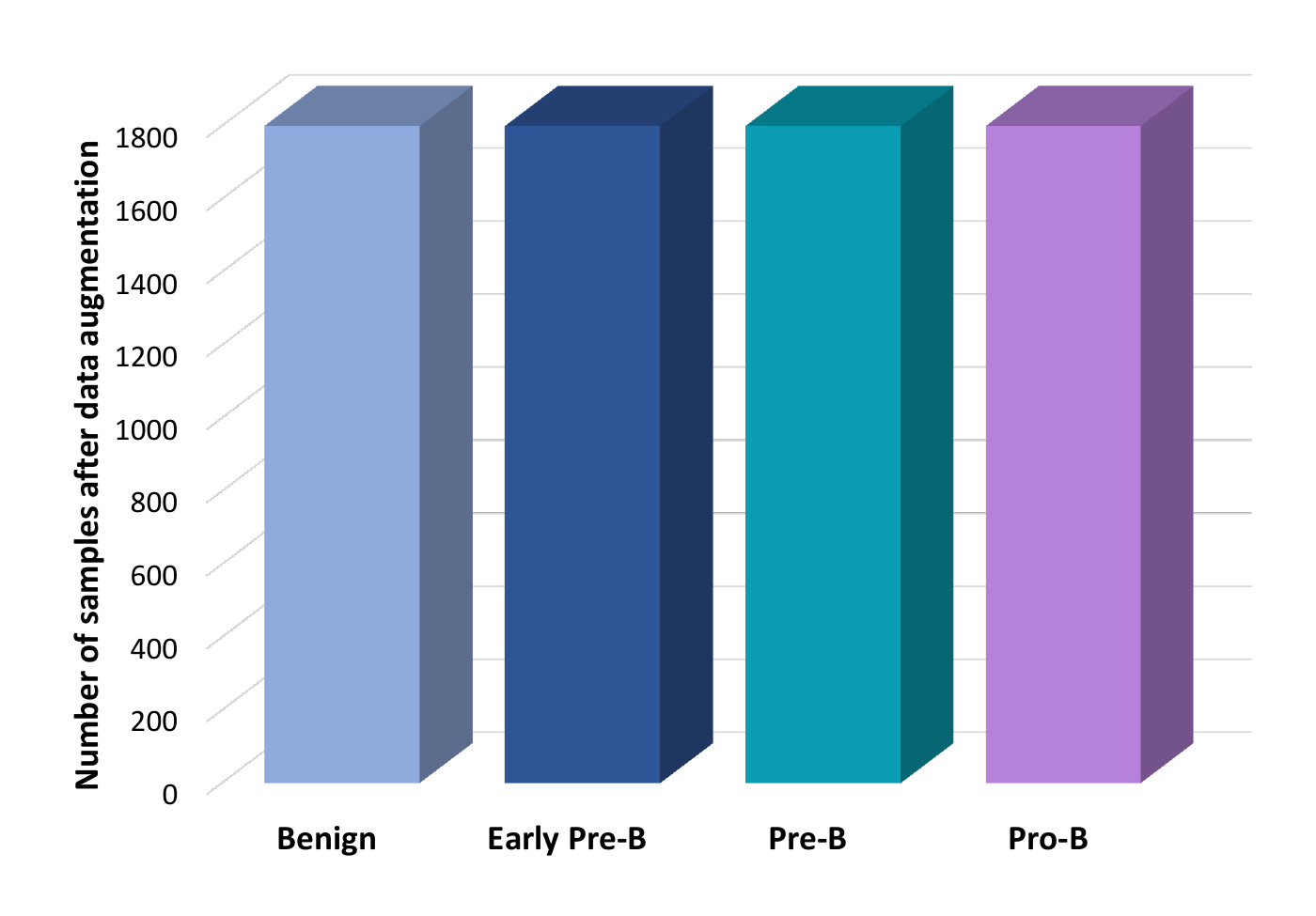}
        
    \end{minipage}
    \caption{(a) Training, Validation, and Testing samples; (b) Number of benign and malignant cells.}
    \label{fig:side_by_side}
\end{figure}
\begin{figure}[h]
	\centering
	\includegraphics[width=\textwidth]{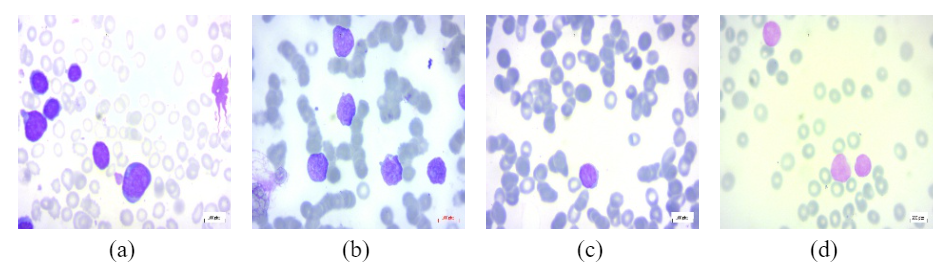}
	\caption{ (a) Benign cell, (b) Early Pre-B cell, (c) Pre-B cell \& (d) Pro-B cell}
\end{figure}

The dataset distinguishes between benign and malignant cells. Benign cells, such as hematogenes, are non-cancerous and generally do not exhibit invasive growth. Although they can cause discomfort or complications by pressing on neighboring structures, they usually pose minimal health risks and rarely recur after treatment.

In contrast, malignant cells are cancerous lymphoblasts characterized by uncontrolled growth and potential to metastasize. These cells can crowd out normal blood cells in the bone marrow and spread to other organs, necessitating prompt diagnosis and targeted treatment. The dataset includes images of four types of malignant lymphoblasts: 979 Early Pre-B, 955 Pre-B, and 796 Pro-B cell images, alongside 512 benign cell images. From these images we separate 100 images from each class and create a test dataset. After combining the prepossess images and original images, we perform data augmentation. This process results in each class having 1800 images. The distribution of these cell types is illustrated in Figure 3 (b), while Figure 4 shows examples of different ALL cell types.
The photographs, captured with a Zeiss camera at 100x magnification and saved as JPG files, were accurately categorized using flow cytometry. The segmented images were obtained through color thresholding in the HSV color space.

\section{Methodology}
This study's methodology is divided into two main sections: image processing of bone marrow smears and deep learning-based ALL cell identification. The first section details the preprocessing pipeline, from raw bone marrow smear images to processed outputs. The following section describes the detection technique, feature extraction, and integration with traditional machine learning classifiers. The mathematical foundations of the technique are also covered. Figure 4.1 illustrates the general workflow.
\begin{figure}[h]
	\centering
	\includegraphics[width=\textwidth]{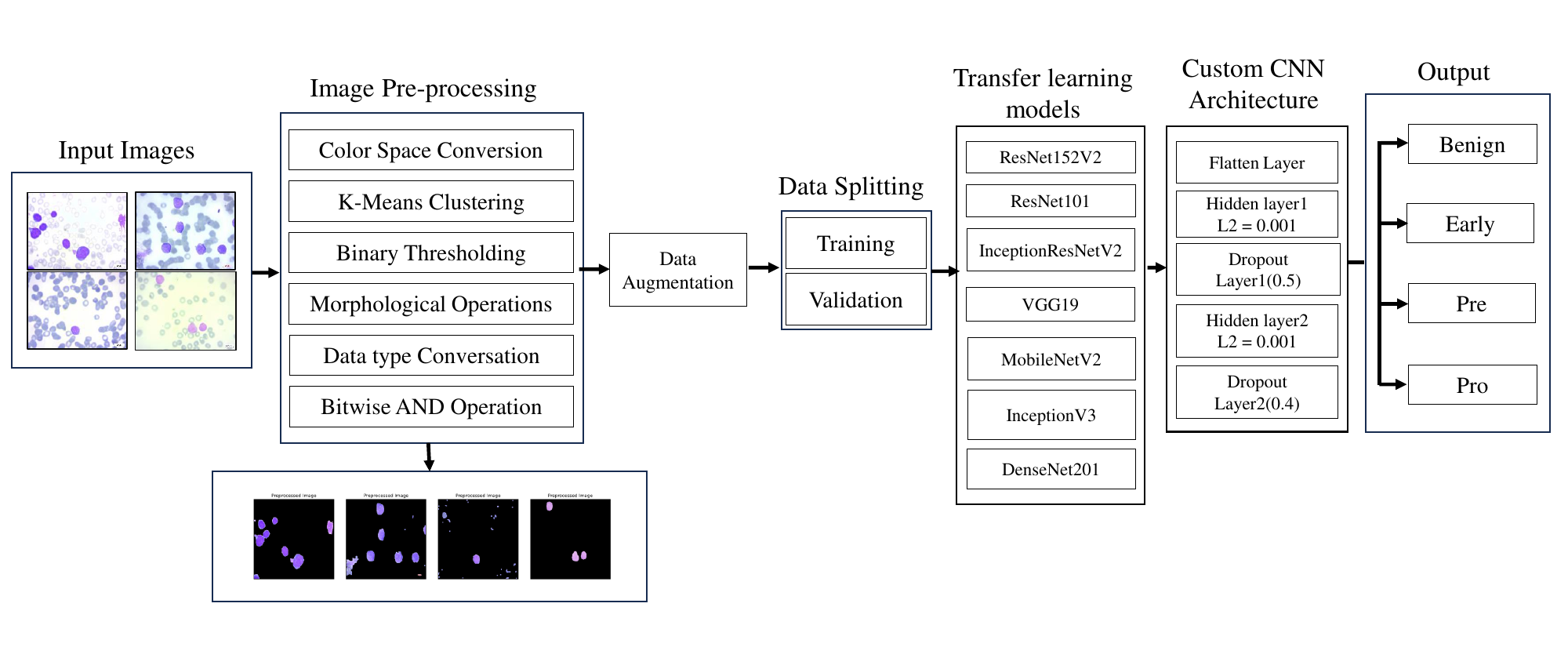}
	\caption{ General flowchart of ALL detection using D-CNN model}
\end{figure}

The system environment setup, including hardware, software, and tools, is outlined. Essential components such as GPUs and specific software libraries are crucial for machine learning tasks. The system specifications for the image processing environment are as follows: it operates on a PC with Windows 10, equipped with an Nvidia 1050 Ti GPU, an Intel Core-i5 2.30 GHz CPU, 8 cores, and 8 GB of memory. For model training, the environment utilizes Kaggle's platform, also running Windows 10. It features a GPU P100, an Intel Xeon 2.30 GHz CPU with 2 cores, and 12 GB of memory.
For deep learning, Nvidia GPUs and cuDNN are used, while Python frameworks and libraries facilitate model training and data visualization, as summarized. The training environment employs TensorFlow v2.10 as the primary framework, with Python as the programming language. Key libraries utilized include NumPy, Pandas, Matplotlib, SciPy, scikit-learn, scikit-image, and OpenCV. These tools collectively support various aspects of model development, data manipulation, and visualization.

\begin{figure}[ht]
	\centering
	\includegraphics[width=0.8\textwidth]{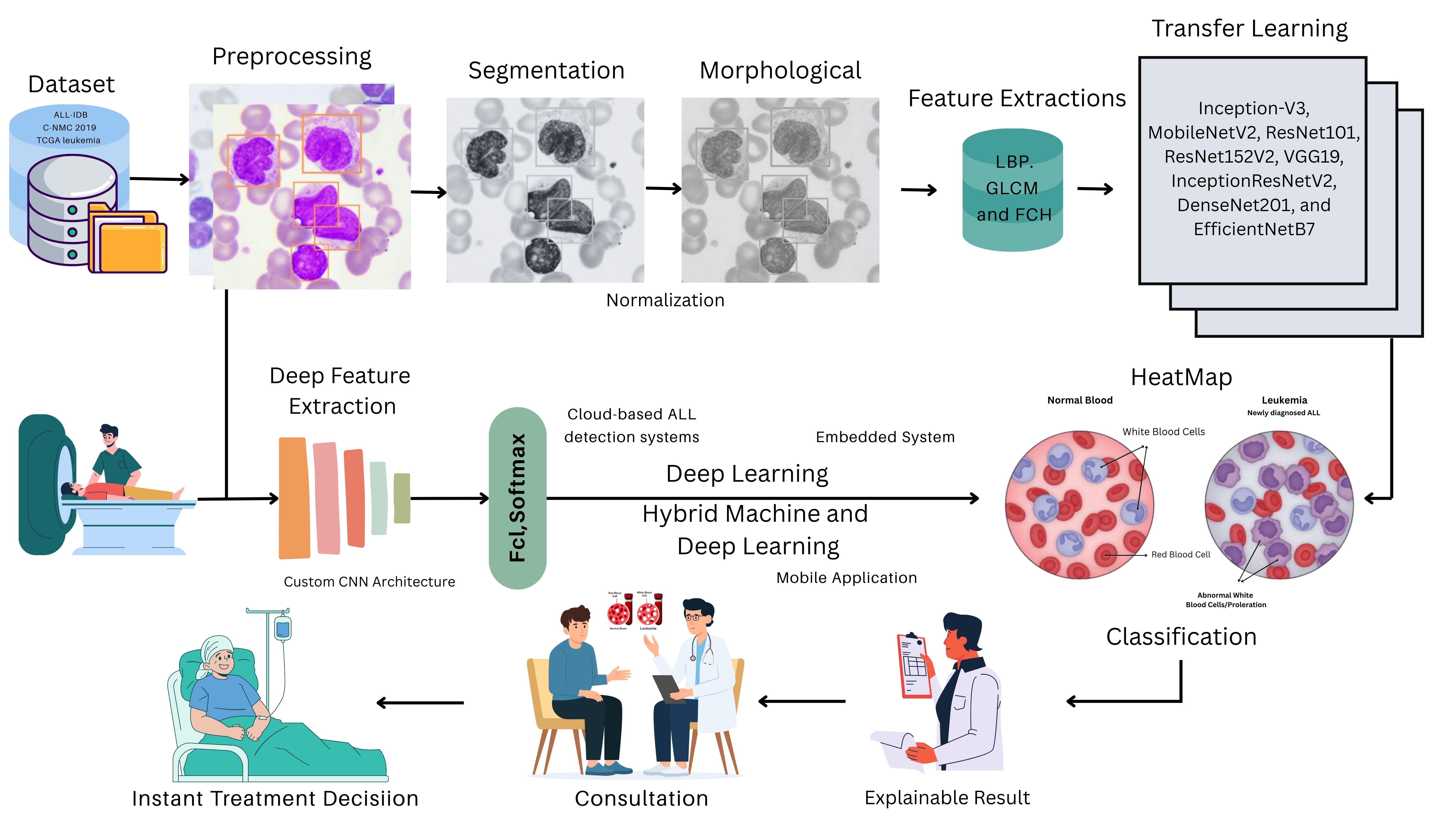}
	\caption{ Methodology of ALL detection}
\end{figure}

The experimental analysis of our methods and results is divided into two sections: ALL cell processing methods \& Innovation and deep neural network model training and classification performance analysis.

\subsection{Image Preprocessing}
Image preprocessing serves as a fundamental step in data analysis and image processing, aimed at enhancing the quality, consistency, and usability of raw data or images for subsequent analysis \cite{maharana2022review}. This crucial phase encompasses a diverse range of techniques, including data cleaning, noise reduction, feature scaling, normalization, and image enhancement. Each of these processes contributes to transforming unrefined data into a standardized and optimized format, thereby facilitating accurate and reliable machine learning and analytical procedures. Ensuring high-quality data preparation is particularly essential when dealing with medical images, where even minor inaccuracies can significantly impact diagnostic performance. Figure 6 presents a structured flowchart depicting the step-by-step approach adopted for image preprocessing in this study.

\begin{figure}[h]
    \centering
    \tikzstyle{startstop} = [rectangle, rounded corners, minimum width=5cm, minimum height=1cm, text centered, draw=black, rotate=90]
    \tikzstyle{arrow} = [thick, ->, >=stealth]
    \begin{tikzpicture}[node distance=2.5cm]
        \node (start) [startstop] at (0, 0) {Input Image};
        \node (res_adj) [startstop, right of=start, xshift = -2.5cm, yshift=-1.5cm] {Resolution Adjustment};
        \node (color_space) [startstop, right of=res_adj, xshift = -2.5cm, yshift=-1.5cm] {Color Space Conversion};
        \node (k_means) [startstop, right of=color_space, xshift = -2.5cm, yshift=-1.5cm] {K-means Clustering};
        \node (binary_thresh) [startstop, right of=k_means, xshift = -2.5cm, yshift=-1.5cm] {Binary Thresholding};
        \node (morph_ops) [startstop, right of=binary_thresh, xshift = -2.5cm, yshift=-1.5cm] {Morphological Operations};
        \node (data_conv) [startstop, right of=morph_ops, xshift = -2.5cm, yshift=-1.5cm] {Data Type Conversion};
        \node (bitwise_and) [startstop, right of=data_conv, xshift = -2.5cm, yshift=-1.5cm] {Bitwise AND Operation};
        \node (output) [startstop, right of=bitwise_and, xshift = -2.5cm, yshift=-1.5cm] {Output Image};
        
        \draw [arrow] (start) -- (res_adj);
        \draw [arrow] (res_adj) -- (color_space);
        \draw [arrow] (color_space) -- (k_means);
        \draw [arrow] (k_means) -- (binary_thresh);
        \draw [arrow] (binary_thresh) -- (morph_ops);
        \draw [arrow] (morph_ops) -- (data_conv);
        \draw [arrow] (data_conv) -- (bitwise_and);
        \draw [arrow] (bitwise_and) -- (output);
    \end{tikzpicture}
    \caption{Image preprocessing approach used in this study}
\end{figure}
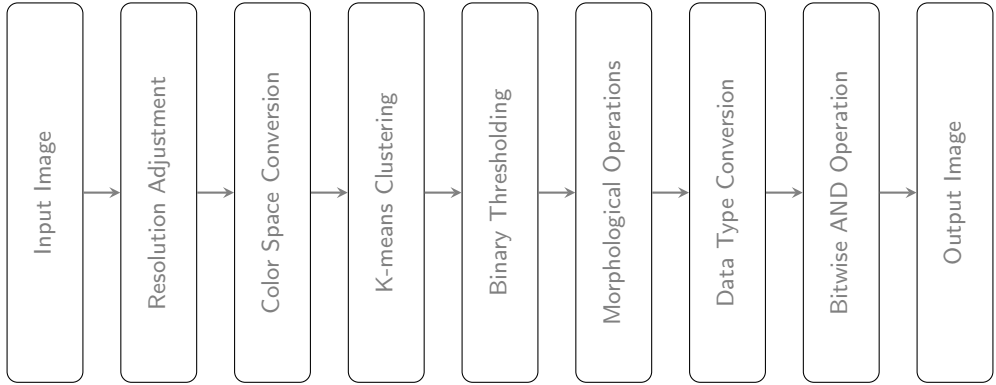

\subsubsection{Resolution Adjustment}
The dataset images were converted to PNG format at a lower resolution, maintaining their aspect ratio. Using the OpenCV library, images were first transformed from BGR to RGB color space. Each image was then resized to 224x224 pixels to meet the input requirements of deep learning models. This process was uniformly applied to all images.
\begin{equation}
I_{\text{resized}} = \text{resize}(I_{\text{input}}, (224,224)) \tag{1}
\end{equation}
\noindent{Where,} \( I_{\text{input}} \) = Original input image, \( I_{\text{resized}} \) = Resized image (224x224 pixels), \texttt{resize()} = Resizing function (e.g., using OpenCV or PIL).

\subsubsection{Color Space Conversion}
Color space conversion involves changing the color representation of an image. We converted RGB images to the LAB color space using \texttt{cv2.cvtColor}, which separates the image into three channels: L (lightness), a* (green to magenta), and b* (blue to yellow). The a* channel was extracted and reshaped into a one-dimensional array for further processing.
\begin{equation}
I_{\text{LAB}} = \text{cv2.cvtColor}(I_{\text{RGB}}, \text{cv2.COLOR\_RGB2LAB}) \tag{2}
\end{equation}
\noindent{Where,} \( I_{\text{RGB}} \) = Original image in RGB color space, \( I_{\text{LAB}} \) = Converted image in LAB color space. The LAB image is split into three channels: L (Lightness), A (Green-Red), and B (Blue-Yellow).

\begin{equation}
I_{L}, I_{A}, I_{B} = \text{split}(I_{\text{LAB}}) \tag{3}
\end{equation}

\subsubsection{K-Means Clustering}
K-Means clustering is an unsupervised method that partitions data into clusters based on proximity to cluster centroids. We applied K-Means to the A channel of the LAB image to group similar colors. The \texttt{KMeans} class from scikit-learn, with \texttt{n\_clusters=7}, was used to find seven color clusters. The clustered A channel was reconstructed and converted to an 8-bit format.
\begin{equation}
\arg \min_{C} \sum_{i=1}^{k} \sum_{x \in C_i} \| x - \mu_i \|^2 \tag{4}
\end{equation}
\noindent{Where, \( k \) = Number of clusters (e.g., 7), \( C_i \) = Set of points in cluster \( i \), \( x \) = Data point in the A channel, \( \mu_i \) = Centroid of cluster \( i \).}

Finally, the clustering result is converted to an 8-bit image:
\begin{equation}
I_{\text{clustered}} = \text{astype}(I_A, \text{uint8}) \tag{5}
\end{equation}

\subsubsection{Binary Thresholding}
Binary thresholding converts a grayscale image into a binary image by classifying pixels as either black (0) or white (1) based on a threshold value. We used \texttt{cv2.threshold} to create a binary image where pixels below the threshold are black and those above are white.
\begin{equation}
I_{\text{binary}}(x, y) = 
\begin{cases} 
255 & \text{if } I_{\text{gray}}(x, y) > T \\
0 & \text{otherwise}
\end{cases} \tag{6}
\end{equation}
\noindent{Where, \( I_{\text{gray}} \) = Grayscale image (A channel of the LAB image), \( T \) = Threshold value (manually or automatically determined).}

\subsubsection{Morphological Operations}
Morphological operations manipulate images based on their shapes. We applied dilation and erosion using functions from the \texttt{scipy.ndimage} module. Dilation expands foreground regions, while erosion reduces them. We used \texttt{binary\_fill\_holes} to fill gaps, \texttt{binary\_opening} to remove noise, and \texttt{binary\_closing} to close gaps in the binary image.

a) Dilation (Expands Regions):
\begin{equation}
I_{\text{dilated}} = I_{\text{binary}} \oplus K = \max_{(s,t) \in K} I_{\text{binary}}(x + s, y + t) \tag{7}
\end{equation}

b) Erosion (Shrinks Regions):
\begin{equation}
I_{\text{eroded}} = I_{\text{binary}} \ominus K = \min_{(s,t) \in K} I_{\text{binary}}(x + s, y + t) \tag{8}
\end{equation}

c) Opening (Erosion Followed by Dilation):
\begin{equation}
I_{\text{opened}} = (I_{\text{binary}} \ominus K) \oplus K \tag{9}
\end{equation}

d) Closing (Dilation Followed by Erosion):
\begin{equation}
I_{\text{closed}} = (I_{\text{binary}} \oplus K) \ominus K \tag{10}
\end{equation}

Where: \( K \) = Structuring element (e.g., a square or circular kernel), \( (x, y) \) = Pixel coordinates, \( (s, t) \) = Offset within the kernel.

\subsubsection{Data Type Conversion}
The binary image was converted to an unsigned 8-bit format. We then used \texttt{cv2.bitwise\_and} to apply a bitwise AND operation between the binary image and the original RGB image. This operation highlights the areas of interest in the original image, which were emphasized during preprocessing.
\begin{equation}
I_{\text{converted}} = \text{astype}(I_{\text{binary}}, \text{uint8}) \tag{11}
\end{equation}

\subsubsection{Bitwise AND Operation}
The bitwise AND operation combines pixel values from two images using a binary mask. The \texttt{selected\_image} is the original image, and \texttt{processed\_image} is the mask with binary values. This operation preserves pixels from the original image where the mask is 1 and sets pixels to 0 where the mask is 0. This technique is useful for object segmentation and targeted image effects.
\begin{equation}
I_{\text{output}} = I_{\text{original}} \land I_{\text{mask}} \tag{12}
\end{equation}
Where: \( I_{\text{original}} \) = Original RGB image, \( I_{\text{mask}} \) = Processed binary mask, \( \land \) = Pixel-wise AND operation (using functions like \texttt{cv2.bitwise\_and}).

\section{ALL Cell Classification Strategy}
Pre-trained models are extensively used to identify Acute Lymphoblastic Leukemia (ALL) cells from medical images. Initially trained on large datasets for general image recognition, these models are subsequently fine-tuned for detecting ALL cells. The training process enables the model to recognize cell morphology and specific features in blood smear or bone marrow images. Once trained, the model can classify images into ALL cell or non-ALL cell categories, assign probability scores, and detect the presence of ALL cells. Post-processing operations can further enhance the results. These models are instrumental in accelerating and improving the accuracy of leukemia diagnosis, aiding in early detection and treatment planning.

\subsection{Convolutional Neural Networks}

Convolutional Neural Networks (CNNs) are specialized deep learning models designed for image and spatial data processing, achieving notable success in computer vision \cite{bhatt2021cnn}. The CNN architecture begins with an image input layer that processes raw pixel values, followed by convolution layers that use learnable filters to generate feature maps \cite{aggarwal2019attention}. A Rectified Linear Unit (ReLU) activation function introduces non-linearity by zeroing negative values \cite{abdallah20221}, while pooling layers, such as max pooling, downsample feature maps and reduce dimensionality \cite{shen2018transdisciplinary}. Fully connected layers then link all neurons in one layer to the next, enabling high-level feature learning, and a softmax layer at the end converts outputs into probability scores for classification tasks. Each component of CNNs is designed to process and interpret image data effectively, making them well-suited for diverse applications.
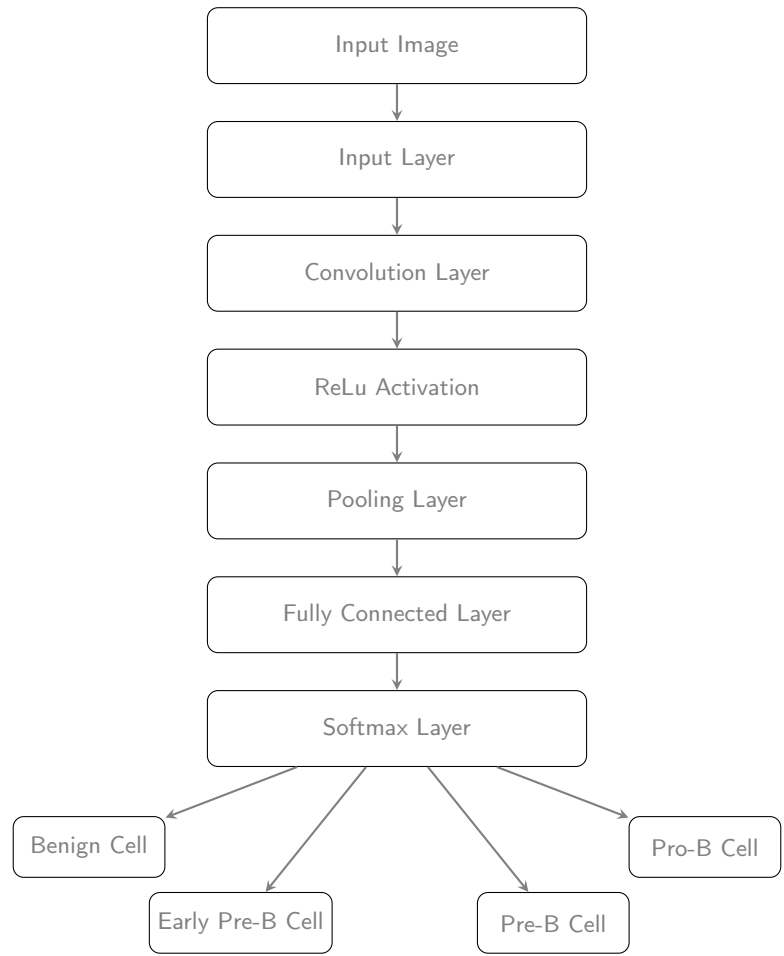
\begin{figure}[h]
	\centering
	\begin{tikzpicture}[node distance=1.5cm]
		
		\tikzstyle{startstop} = [rectangle, rounded corners, minimum width=5cm, minimum height=1cm,text centered, draw=black]
		\tikzstyle{cell} = [rectangle, rounded corners, minimum width=2cm, minimum height=0.8cm,text centered, draw=black]
		\tikzstyle{arrow} = [thick,->,>=stealth]
		
		\node (start) [startstop] at (0,0) {Input Image};
		\node (a) [startstop, below of=start] {Input Layer};
		\node (b) [startstop, below of=a] {Convolution Layer};
		\node (c) [startstop, below of=b] {ReLu Activation};
		\node (d) [startstop, below of=c] {Pooling Layer};
		\node (e) [startstop, below of=d] {Fully Connected Layer};
		\node (f) [startstop, below of=e] {Softmax Layer};
		\node (g) [cell, below left of=f, yshift = -0.5cm, xshift=-3cm] {Benign Cell};
		\node (h) [cell, below left of=f, yshift = -1.5cm, xshift=-1cm] {Early Pre-B Cell};
		\node (i) [cell, below right of=f, yshift = -1.5cm, xshift=1cm] {Pre-B Cell};
		\node (j) [cell, below right of=f, yshift = -0.5cm, xshift=3cm] {Pro-B Cell};
		
		\draw [arrow] (start) -- (a);
		\draw [arrow] (a) -- (b);
		\draw [arrow] (b) -- (c);
		\draw [arrow] (c) -- (d);
		\draw [arrow] (d) -- (e);
		\draw [arrow] (e) -- (f);
		\draw [arrow] (f) -- (g);
		\draw [arrow] (f) -- (h);
		\draw [arrow] (f) -- (i);
		\draw [arrow] (f) -- (j);
		
	\end{tikzpicture}
	\caption{Architecture of Convolutional Neural Network}
\end{figure}
\subsection{Transfer Learning with Pre-trained Models}
Transfer learning utilizes pre-trained models to leverage the feature representations learned from large datasets, providing a head start in identifying patterns and features for new tasks \cite{torrey2010transfer}. This technique accelerates model development, reduces the need for extensive labeled data, and often improves performance by fine-tuning pre-trained models to specific tasks. The process involves selecting a suitable pre-trained model, adapting its architecture or layers, and training it on a new dataset. This approach, central to deep learning, allows for the creation of complex models with fewer computations and faster convergence, revolutionizing various applications from image recognition to natural language processing. For this study, we evaluated several pre-trained models, including Inception-V3, MobileNetV2, ResNet101, ResNet152V2, VGG19, InceptionResNetV2, DenseNet201, and EfficientNetB7, to identify the most effective feature extractor for our dataset. The general equation for a convolutional neural network (CNN)-based model, such as Inception-V3, MobileNetV2, ResNet101, etc. can be written as:

\begin{equation}
\text{Model}(x) = \text{Softmax}\left(\text{FullyConnected}\left(\mathcal{F}(x)\right)\right)
\end{equation}

\text{Where: } \( x \) = Input image (usually preprocessed, such as resized, normalized, etc.), \( \mathcal{F}(x) \) = Feature extraction process (the series of convolutional, pooling, activation, and/or residual layers specific to each model), \( \text{FullyConnected} \) = The fully connected layer that connects the extracted features to the output, \( \text{Softmax} \) = The final activation function that outputs class probabilities (in classification tasks).

\subsubsection{InceptionResNetV2 (Hybrid Architecture)}

InceptionResNetV2 is a powerful hybrid architecture that skillfully merges the multi-scale processing capabilities of Inception modules with the robust training characteristics of residual connections. Its core idea revolves around combining the strengths of both Inception and ResNet architectures by using Inception-style modules but crucially incorporating residual connections within them. This means that each Inception module's output is added to its input (after a 1x1 convolution for channel alignment if needed), a process often facilitated by a "Filter Expansion Layer" which is a 1x1 convolution without activation used to scale up the dimensionality before the residual addition. With 164 layers, this model excels in image pattern recognition, as the integration of Inception modules with residual connections serves to stabilize training and significantly reduce vanishing gradients. Further architectural features like reduction blocks and global average pooling facilitate efficient dimensionality reduction. While generally more computationally expensive than other Inception versions due to its depth and complexity, InceptionResNetV2 consistently achieves very high accuracy, leveraging its combined strengths to excel in image classification, object detection, and segmentation tasks, especially when pre-trained on large datasets like ImageNet.

\subsubsection{ResNet 101 \& ResNet152V2 (Residual Networks)}

ResNet-101, a deep CNN architecture from the Residual Network family, is known for its depth and ability to learn complex hierarchical features from images \cite{singh2021frequency}. With 101 layers, it effectively handles challenging computer vision tasks through the use of residual blocks, which introduce skip connections that facilitate training deep networks \cite{oyedotun2021training}. The network utilizes bottleneck residual blocks for computational efficiency while maintaining depth \cite{queiruga2020continuous}. ResNet-101 is typically pre-trained on datasets like ImageNet, providing useful feature representations that can be fine-tuned for specific tasks. The architecture's skip connections support gradient flow, addressing the vanishing gradient problem \cite{reena2022content}. It uses global average pooling in the final layers to reduce feature maps to fixed-sized vectors \cite{zhang2019pyramid}. The network aims to fit residual mappings rather than the original ones:
\begin{equation}
F(x) = H(x) - x
\end{equation}

The residual mapping is then recast as:
\begin{equation}
H(x) = F(x) - x
\end{equation}

\subsubsection{MobileNetV2 (Mobile-Optimized Architecture)}

MobileNetV2 advances CNN architecture for mobile and embedded applications, improving on MobileNet with "inverted residual blocks" \cite{tu2020pruning}. These blocks feature depthwise separable convolutions, which split the convolution process into depthwise and pointwise steps, reducing computational cost \cite{wang2020new}. The linear bottleneck layer optimizes feature channel compression \cite{singh2019shunt}. MobileNetV2 includes hyperparameters like the width multiplier ($\alpha$) and resolution multiplier ($\rho$), which adjust feature channels and input image size, respectively \cite{janarthan2022p2op}. Its design emphasizes efficiency and speed, making it suitable for mobile and embedded devices with limited resources \cite{li2022mobilenetv2}.

\subsubsection{DenseNet201 (Densely Connected Convolutional Networks)}

DenseNet201, with 201 layers, features dense connectivity where each layer receives inputs from all previous layers, promoting feature reuse and mitigating vanishing gradients. This structure enhances training and generalization. Pre-trained on datasets like ImageNet, DenseNet201 excels in various computer vision applications due to its robustness and efficient parameter use.

\subsubsection{InceptionV3 (GoogleNet/Inception Architecture) }
Inception-V3, or GoogLeNetV3, represents a significant advancement in deep convolutional neural networks, designed for image classification and object recognition \cite{padilla2020deep}. It builds on the original GoogLeNet with enhanced "Inception" modules and a deep architecture comprising 48 layers \cite{jena2022convolutional}. These modules include parallel convolutional layers of various sizes (1x1, 3x3, 5x5) and integrate pooling and dimensionality reduction, which aids in capturing complex patterns and features at multiple scales \cite{chantrapornchai2023micro}. Batch normalization throughout the network improves stability and efficiency. Inception-V3 also employs factorized convolutions, breaking down conventional convolutions into 1xN and Nx1 steps to optimize parameter count and processing load \cite{shubha2022image}. Auxiliary classifiers address the vanishing gradient problem, enhancing overall performance \cite{shyamalee2022cnn}. Typically pre-trained on large datasets like ImageNet, Inception-V3 learns extensive feature representations, which are then fine-tuned for specific tasks \cite{bukhsh2021damage}. Its efficient design balances model size and accuracy, making it suitable for real-time applications and limited-resource scenarios. Inception-V3's contributions continue to impact deep learning, excelling in image-related tasks and transfer learning scenarios \cite{huang2023towards}.
\subsubsection{VGG19 (Visual Geometry Group)}

VGG19 is noted for its simplicity and effectiveness, consisting of 19 layers with 16 convolutional layers and 3 fully connected layers. It uses 3x3 convolutional kernels and max-pooling layers, making it effective for image recognition and object classification.
Despite its simpler design compared to more recent architectures, VGG19 remains competitive and valuable, particularly for pretraining on large datasets and fine-tuning for specific tasks.
\begin{landscape}
\begin{table}[ht]
    \centering
    \caption{Comprehensive Comparison of CNN Architectures}
    \vspace{1em}
    \small
    \setlength{\tabcolsep}{4pt} 
    \begin{tabular}{p{2cm} p{3cm} p{3cm} p{2cm} p{1.2cm} p{3cm} p{3cm} p{4cm}}
        \hline
        \textbf{Model} & \textbf{Architecture Type \& Design Philosophy} & \textbf{Core Innovations \& Blocks} & \textbf{Depth \& Parameters} & \textbf{FLOPs (GFLOPs)} & \textbf{Training Behavior \& Optimization} & \textbf{Strengths \& Advantages} & \textbf{Limitations \& Typical Use Cases} \\
        \hline
        
        VGG19 & Plain, uniform CNN stacking small 3×3 convolutions & Sequential blocks of 3×3 conv + ReLU; max-pooling after conv blocks & 19 layers; $\sim$143M parameters & $\sim$19.6 & Easy to optimize; prone to overfitting; large memory/computation due to FC layers & Simple, interpretable; strong hierarchical feature extraction; widely used baseline & Large size and compute; slow training/inference; lacks multi-scale features; baseline, transfer learning, education \\
        \hline
        
        InceptionV3 & Multi-path inception modules enabling multi-scale feature extraction & Parallel convs (1×1, 3×3, 5×5), factorized convolutions, auxiliary classifiers & $\sim$159 layers (modules); $\sim$24M parameters & $\sim$5.7 & Factorization reduces compute; stable training with auxiliary classifiers & Efficient multi-scale features; reduced parameters and compute; good accuracy & Complex design; harder to implement; image classification, detection; balance accuracy-efficiency \\
        \hline
        
        ResNet101 & Deep residual network addressing vanishing gradients & Residual blocks with identity skip connections; bottleneck convs for efficiency & 101 layers; $\sim$44.5M parameters & $\sim$7.8 & Skip connections ease very deep training; batch norm + SGD with momentum & Enables very deep networks; improved accuracy and gradient flow & Longer training; higher memory; complexity due to residuals; classification, segmentation \\
        \hline
        
        ResNet152V2 & Pre-activation ResNet improving gradient flow & Pre-activation residual blocks (BN-ReLU-Conv order), bottleneck design & 152 layers; $\sim$60M parameters & $\sim$11.3 & Pre-activation stabilizes training; better convergence & More stable training of deeper nets; improved accuracy over original ResNet & Higher compute and memory; training complexity; large-scale recognition, fine-grained classification \\
        \hline
        
        Inception-
        ResNetV2 & Hybrid of Inception and Residual networks combining benefits & Residual inception modules; 1×1 conv expansion filters; hybrid design & $\sim$572 layers (module basis); $\sim$56M parameters & $\sim$13.2 & Combines factorized convs with residual learning; careful learning rate/regularization needed & State-of-the-art accuracy; multi-scale + residual benefits; smooth gradient flow & Computationally expensive; complex design; longer training; cutting-edge classification and research \\
        \hline
        
        MobileNetV2 & Mobile-optimized, lightweight CNN for embedded devices & Depthwise separable convs; inverted residual blocks with linear bottlenecks & $\sim$88 layers; $\sim$3.4M parameters & $\sim$0.3 & Designed for fast convergence on mobile; uses ReLU6 activations, batch norm & Very low compute; fast inference; small model size; good accuracy on mobile & Sacrifices some accuracy; limited capacity for complex features; mobile/IoT/real-time apps \\
        \hline
        
        DenseNet201 & Dense connectivity maximizing feature reuse and gradients & Dense blocks where every layer connects to all subsequent layers; transitions reduce complexity & 201 layers; $\sim$20M parameters & $\sim$4.3 & Strong gradient flow; parameter efficient; batch norm + SGD momentum & High parameter efficiency; strong feature propagation; smaller than similar depth nets & Memory overhead due to concatenations; complex graph; research, memory-limited tasks \\
        \hline
    \end{tabular}
\end{table}
\end{landscape}

We began our research by outlining the tools used, including both hardware and software, and their contributions to the study. Next, we detailed the preprocessing phase, which involved improving image quality, ensuring consistency, and minimizing noise to prepare the medical images for analysis. These preprocessing steps were vital for ensuring the suitability of the images for further examination. Finally, we concentrated on the core of our research—the Convolutional Neural Network (CNN) models. We described the selection and customization of these CNNs to align with our research goals, including their design, structure, and training methodologies, to provide a comprehensive understanding of their role in our study.

\subsection{Innovation}
\subsubsection{Multi-Head Self-Attention Mechanism}
This article introduces a multi-head self-attention mechanism into the model. The core idea is to map the input into multiple subspaces through multiple heads, perform attention computations independently within each subspace, and then concatenate the results from all subspaces. This enhances the model’s feature representation capability.

The multi-head self-attention mechanism first linearly transforms the input feature representations (in matrix form) into three spaces: Query (Q), Key (K), and Value (V). Then, in each subspace, it computes attention scores by taking the dot product of the queries and keys, applies these weights to the values, and obtains the attention output.

The specific formula for the attention mechanism is as follows:
\begin{equation}
\text{Attention}(Q, K, V) = \text{softmax}\left(\frac{QK^T}{\sqrt{d_k}}\right) V
\end{equation}

Where:
\begin{itemize}
    \item $Q$, $K$, and $V$ are the \textbf{query}, \textbf{key}, and \textbf{value} matrices, respectively.
    \item $d_k$ is the dimensionality of the key matrix.
\end{itemize}

The computation process of multi-head self-attention involves several key steps — Linear Transformation, Dot-Product Similarity, Softmax, Weighted Sum, and Multi-Head Concatenation. First, the input features are linearly transformed to generate the query ($Q$), key ($K$), and value ($V$) matrices, allowing the model to project the input into multiple subspaces. Next, the dot product between $Q$ and $K$ is calculated and scaled by $\sqrt{d_k}$ to obtain similarity scores. These scores are then passed through a softmax function to generate attention weights, indicating the relative importance of elements in the sequence. The attention weights are applied to $V$ to compute the weighted sum, producing the attention output. Finally, the outputs from all attention heads are concatenated to form the final representation. This multi-head approach enables the model to capture rich, diverse global dependencies, enhancing its feature representation capability.

\subsubsection{Focal Loss}
To address the issue of class imbalance, this article adopts Focal Loss as the loss function. Focal Loss is an improved version of the cross-entropy loss, specifically designed to tackle class imbalance problems. Class imbalance can cause the model to favor predicting the majority class, thereby neglecting minority class samples. Focal Loss introduces a modulation factor to the standard cross-entropy loss, enabling the model to focus more on hard-to-classify samples during training, thus improving the model's robustness.

The formula for Focal Loss is:
\begin{equation}
FL(p_t) = -\alpha_t (1 - p_t)^\gamma \log(p_t)
\end{equation}

Where:
\begin{itemize}
    \item \( \alpha_t \) is the balancing factor,
    \item \( \gamma \) is the modulation factor (focusing parameter),
    \item \( p_t \) is the model’s predicted probability for the correct class.
\end{itemize}
The balancing factor \( \alpha_t \) helps prevent the loss from being dominated by the majority class by adjusting the contribution from different classes. By setting an appropriate \( \alpha_t \), the model can focus more on minority class samples during training. The modulation factor \( \gamma \) controls the focus on hard-to-classify examples. When \( \gamma = 0 \), Focal Loss reduces to standard cross-entropy. As \( \gamma \) increases, the loss amplifies the contribution of hard samples, which have low predicted probabilities \( p_t \). The predicted probability \( p_t \) reflects the model's confidence in the correct class, and by adjusting its influence, Focal Loss dynamically balances the loss between easy and hard samples. This approach helps mitigate the negative effects of class imbalance by emphasizing difficult examples during training.

\subsection{Model Training and Evaluation}
Initially, we used pre-trained models for transfer learning on our preprocessed dataset. We modified these models by replacing the classifier layer with a Dense layer matching our classification task's number of classes and introducing a Global Average Pooling layer. Early stopping was implemented to prevent overfitting by halting training when validation loss did not improve. Models were initialized with ImageNet pre-trained weights, leveraging diverse features for better performance. The Adam optimizer with a learning rate of 0.0001 was used during training, with hyperparameters detailed in Table 5.1. We employed the sparse categorical cross-entropy loss function, suitable for classification problems with integer target labels.

\begin{figure}[h]
	\centering
	\includegraphics[width=\textwidth]{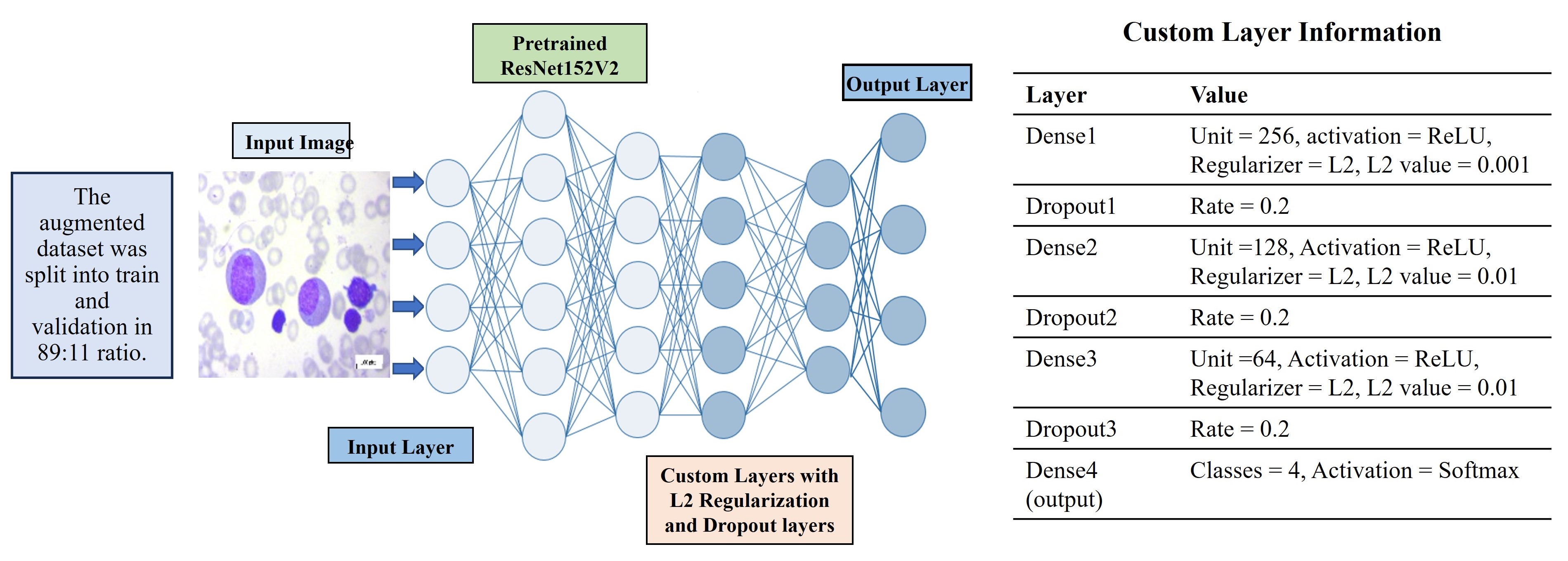}
	\caption{ Model Development and Training}
\end{figure}

\begin{center}

\begin{tabular}{| p{3cm} | p{6cm} |}
    \hline
    \textbf{Parameter}  & \textbf{Value}                           \\ \hline
    Learning Rate       & 0.0001                                   \\ \hline
    Batch Size          & 32                                       \\ \hline
    Epoch               & 30                                       \\ \hline
    Optimizer           & Adaptive Moment Estimation               \\ \hline
    Activation Function & ReLU, SoftMax                            \\ \hline
    Loss Function       & Categorical Cross-entropy, Focal Loss    \\ \hline
    Custom Layer        & 1 Flatten, 2 Hidden, 2 Dropout, 1 Output \\ \hline
    Train Data          & 6480                                     \\ \hline
    Validation Data     & 720                                      \\ \hline
    Test Data           & 400                                      \\ \hline
    Regularizer         & L2                                       \\ \hline
\end{tabular}
\captionof{table}{Hyperparameters used for tuning the networks.}
\end{center}

\subsection{Performance Metrics}
To evaluate the efficacy of the trained models, we computed several performance metrics, including classification accuracy, precision, recall, F1 score, training and validation loss, and Confusion Matrix. These metrics provide insights into the models' performance and their ability to discriminate between classes.

\subsubsection{Accuracy}
Accuracy is the proportion of correctly predicted samples out of all samples. The formula is as follows:
\begin{equation}
\text{Accuracy} = \frac{TP + TN}{TP + FP + TN + FN}
\end{equation}
\noindent where \\
TP = True Positive, TN = True Negative, FP = False Positive, FN = False Negative
These values are also used to calculate precision and recall. 
\subsubsection{Precision}
Precision is the proportion of correctly predicted positive samples among all samples predicted as positive. Precision is calculated as:
\begin{equation}
\text{Precision} = \frac{TP}{TP + FP}
\end{equation}

\subsubsection{Recall}
Recall is the proportion of actual positive samples correctly predicted as positive by the model.  Recall is calculated as:
\begin{equation}
\text{Recall} = \frac{TP}{TP + FN}
\end{equation} 

\subsubsection{F1 Score}
The F1-score is the harmonic mean of precision and recall, used for comprehensive evaluation of the model's classification performance. 
\begin{equation}
F 1 \text { Score }=2 \cdot \frac{\text { Precision } \times \text { Recall }}{\text { Precision }+ \text { Recall }}
\end{equation}

\subsubsection{Loss}
Loss measures the difference between predicted and actual values, indicating how well a model performs.

\subsubsection{Confusion Matrix}
A confusion matrix is a tool used to evaluate the performance of classification models by comparing predicted outcomes to actual results. For binary classification, it consists of four key components: True Positives (TP), True Negatives (TN), False Positives (FP), and False Negatives (FN). This 2x2 table can be expanded for multi-class problems, providing insights into where the model succeeds or fails, and is essential for assessing model accuracy and errors.

\begin{center}

\vspace{0.5em}
\begin{tabular}{|l|l|l|}
    \hline
    \textbf{} & \textbf{Predicted Positive} & \textbf{Predicted Negative} \\ \hline
    \textbf{Actual Positive} & True Positives (TP) & False Negatives (FN) \\ \hline
    \textbf{Actual Negative} & False Positives (FP) & True Negatives (TN) \\ \hline
\end{tabular}
\captionof{table}{Visual representation of confusion matrix for binary classification}
\end{center}

By comparing the evaluation results of different models, we can observe performance differences in the in the detection and classification of ALL cell types. The evaluation metrics used in the project include accuracy, precision, recall, and F1-score, and performance comparisons are presented through bar charts to select the best-performing model.

\begin{figure}[ht]
    \centering
    \begin{subfigure}[b]{0.45\textwidth}
        \centering
        \includegraphics[width=\textwidth]{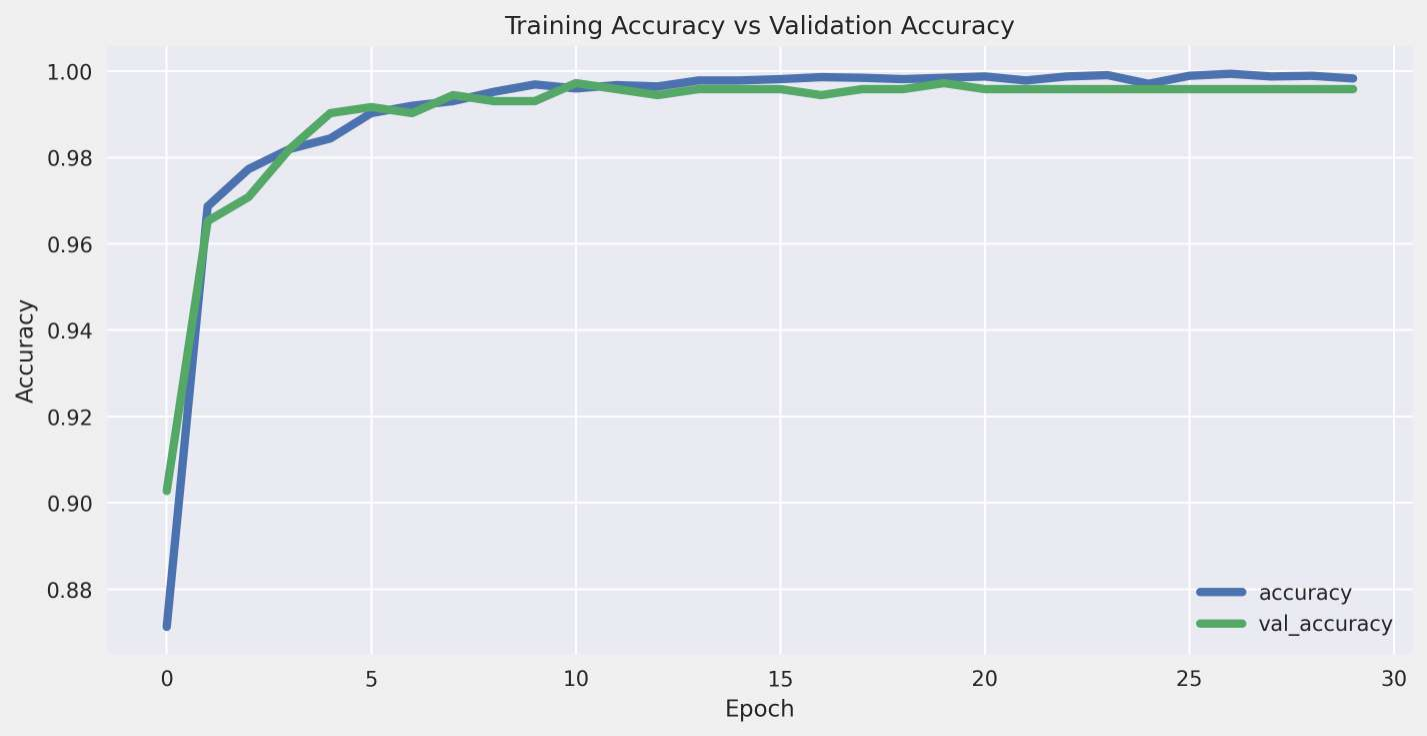}
        \caption{DenseNet 201}
    \end{subfigure}
    \hfill
    \begin{subfigure}[b]{0.45\textwidth}
        \centering
        \includegraphics[width=\textwidth]{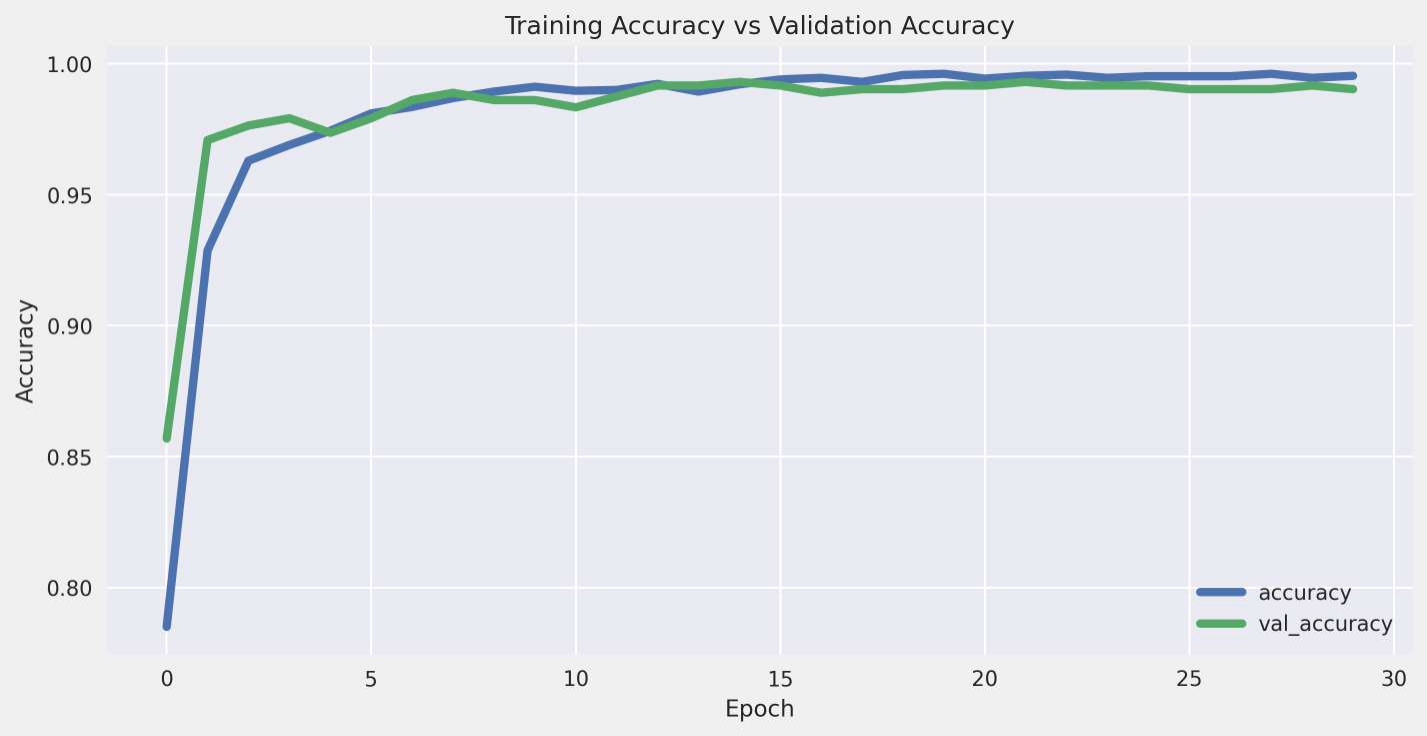}
        \caption{InceptionResNetV2}
    \end{subfigure}
    \vfill
    \begin{subfigure}[b]{0.45\textwidth}
        \centering
        \includegraphics[width=\textwidth]{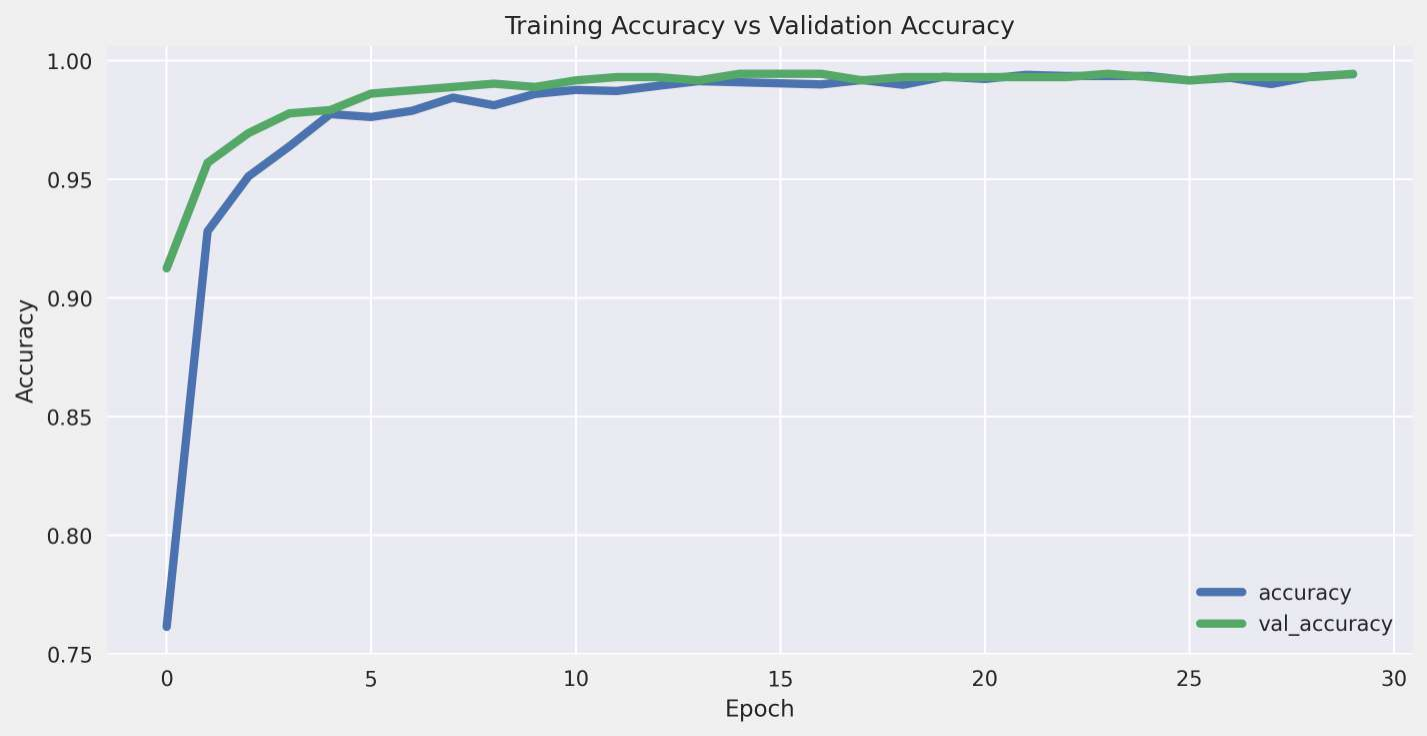}
        \caption{InceptionV3}
    \end{subfigure}
    \vfill
    \begin{subfigure}[b]{0.45\textwidth}
        \centering
        \includegraphics[width=\textwidth]{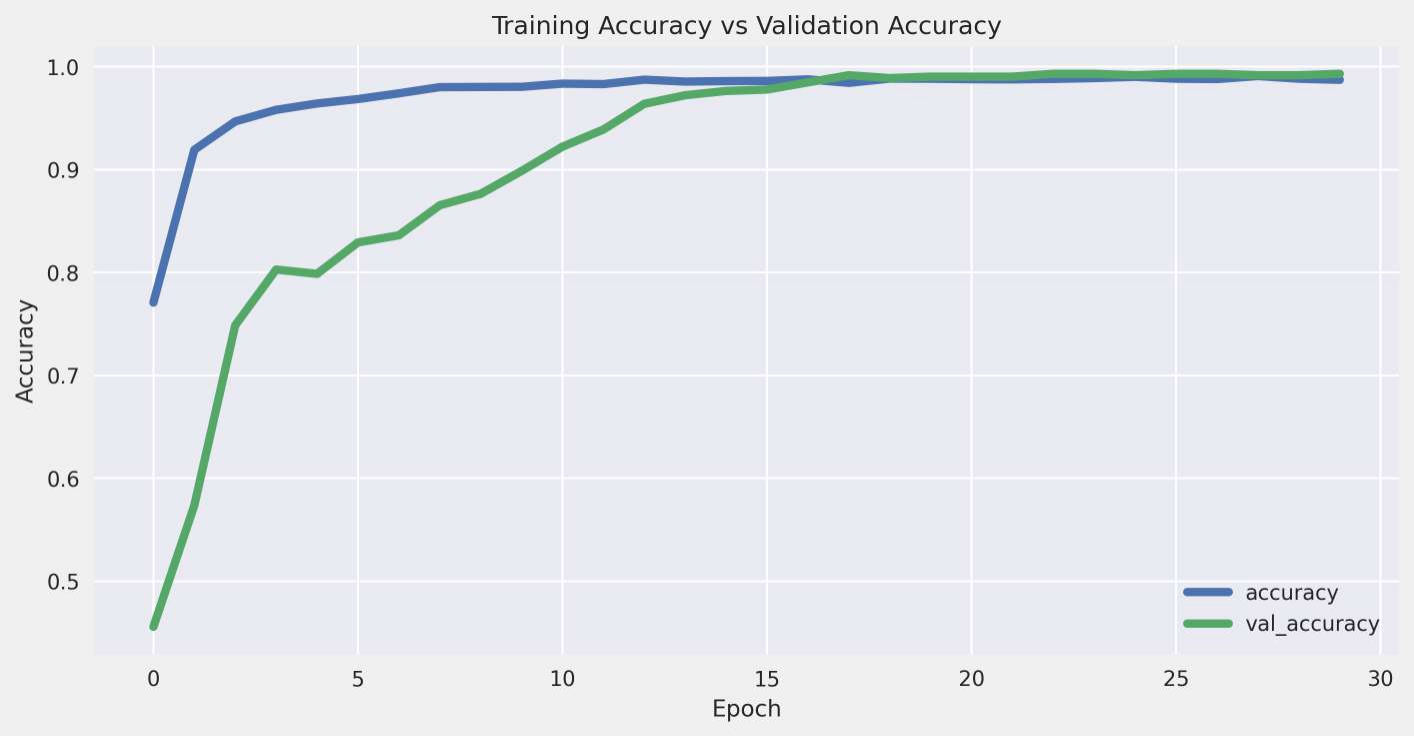}
        \caption{MobileNetV2}
    \end{subfigure}
    \hfill
    \begin{subfigure}[b]{0.45\textwidth}
        \centering
        \includegraphics[width=\textwidth]{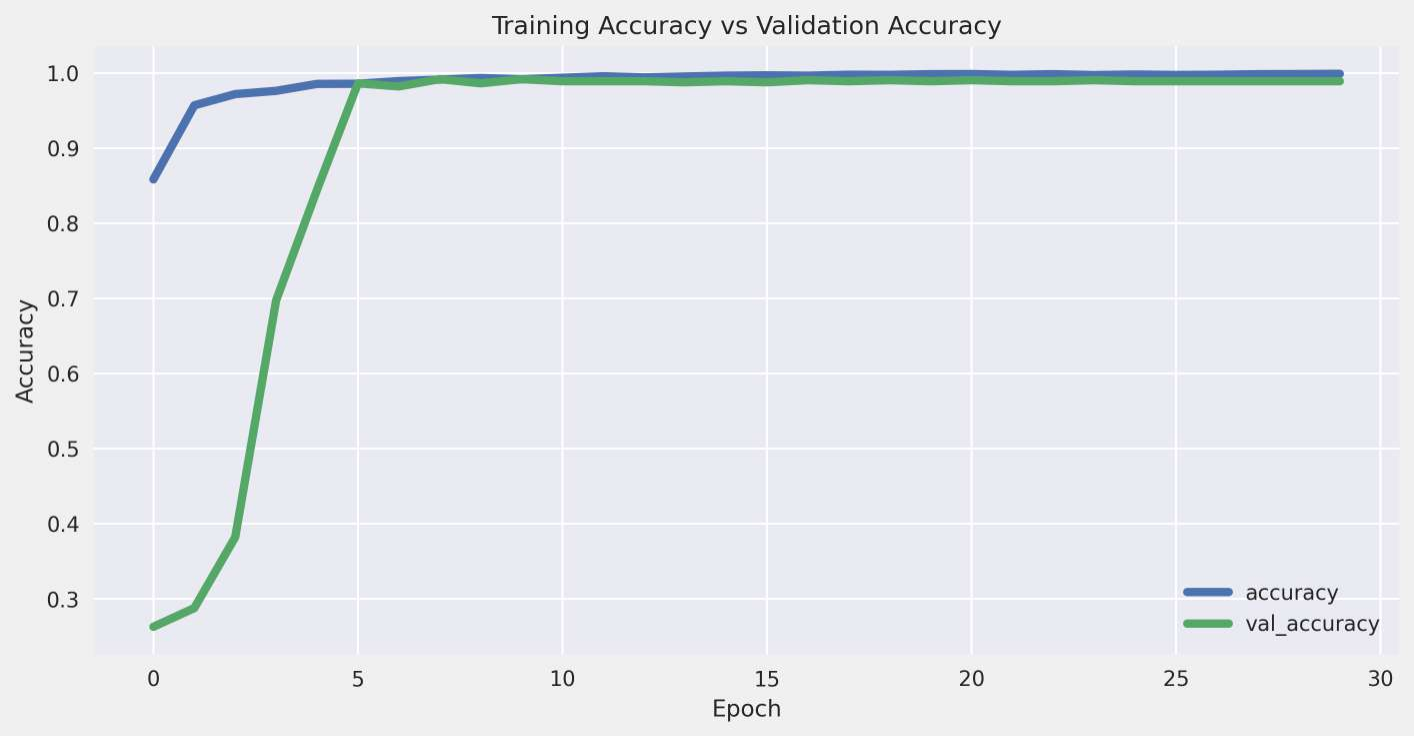}
        \caption{ResNet101}
    \end{subfigure}
    \vfill
    \begin{subfigure}[b]{0.45\textwidth}
        \centering
        \includegraphics[width=\textwidth]{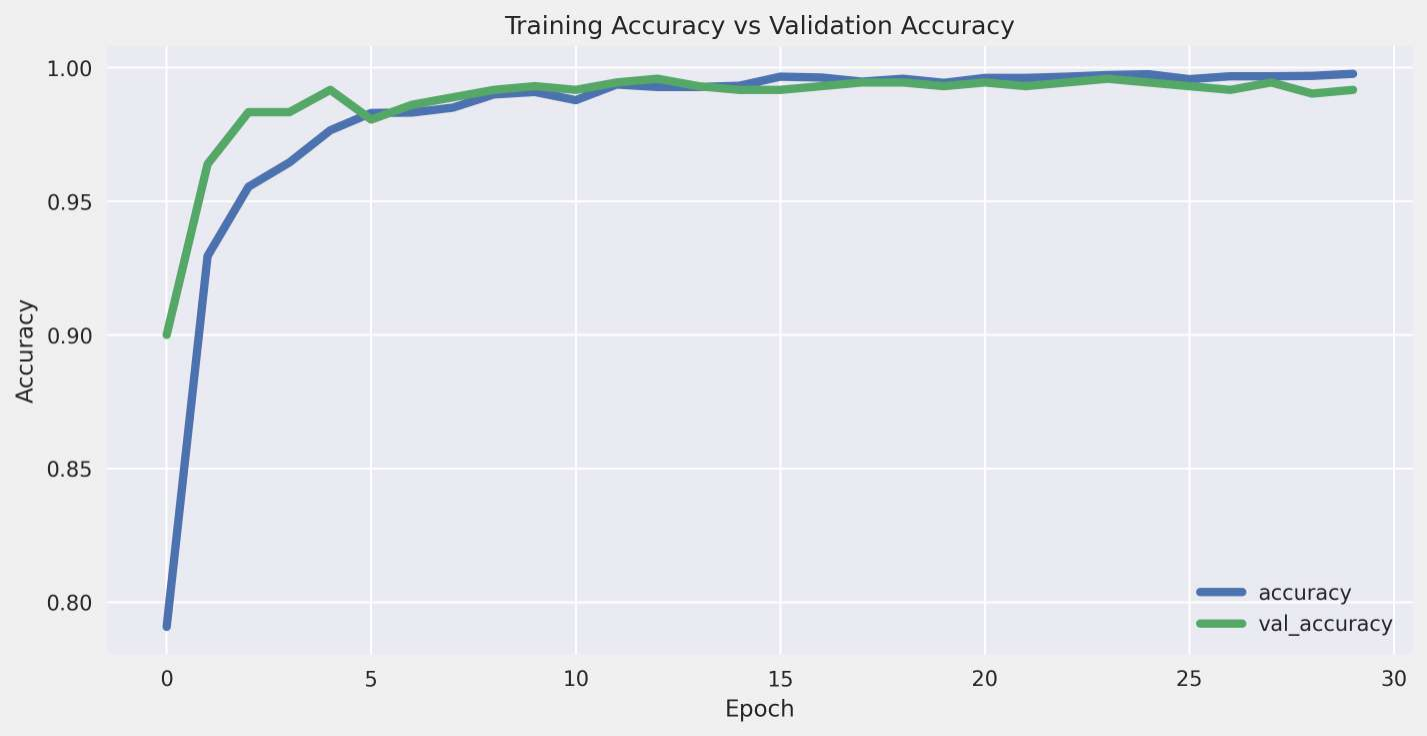}
        \caption{ResNet152 V2}
    \end{subfigure}
    \hfill
    \begin{subfigure}[b]{0.45\textwidth}
        \centering
        \includegraphics[width=\textwidth]{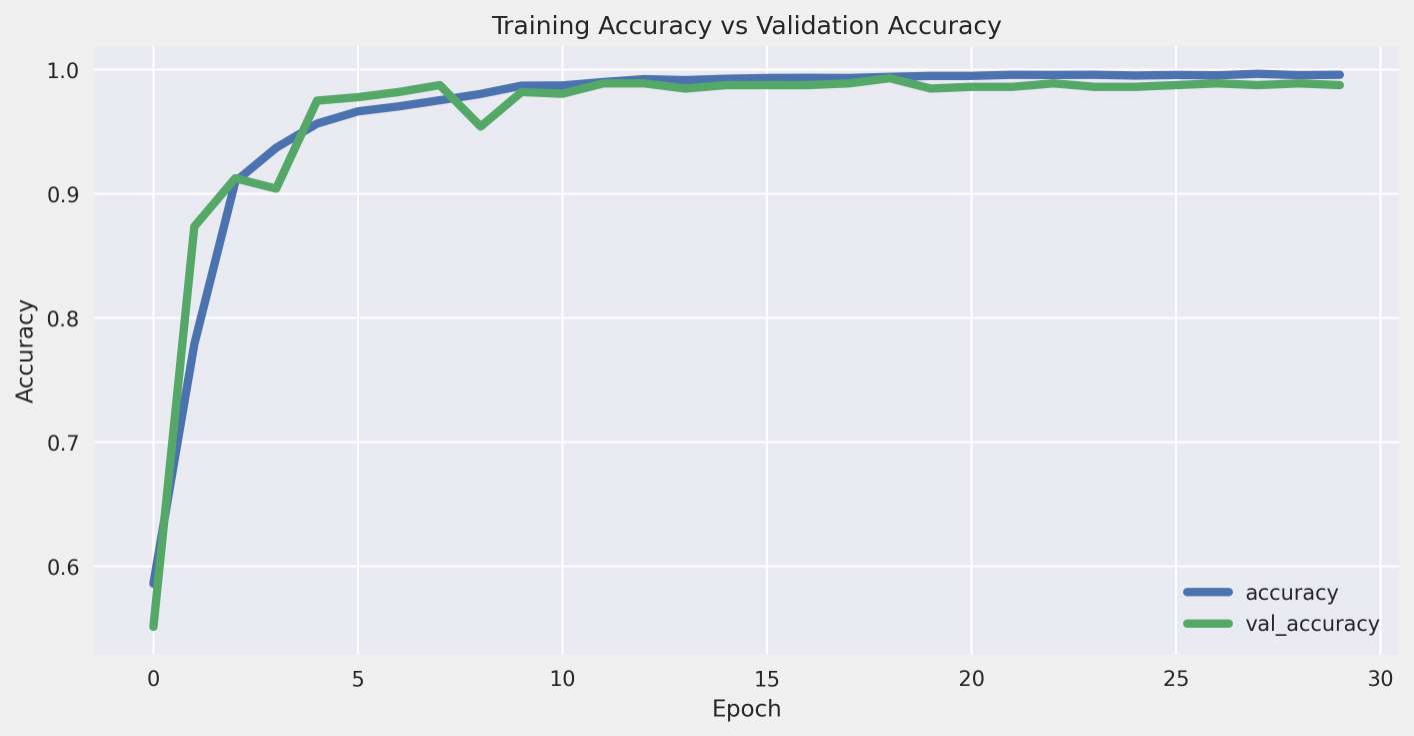}
        \caption{VGG19}
    \end{subfigure}
    \caption{Training accuracy vs validation accuracy graph}
\end{figure}

\clearpage
\begin{figure}[h]
    \centering
    \begin{subfigure}[b]{0.45\textwidth}
        \centering
        \includegraphics[width=\textwidth]{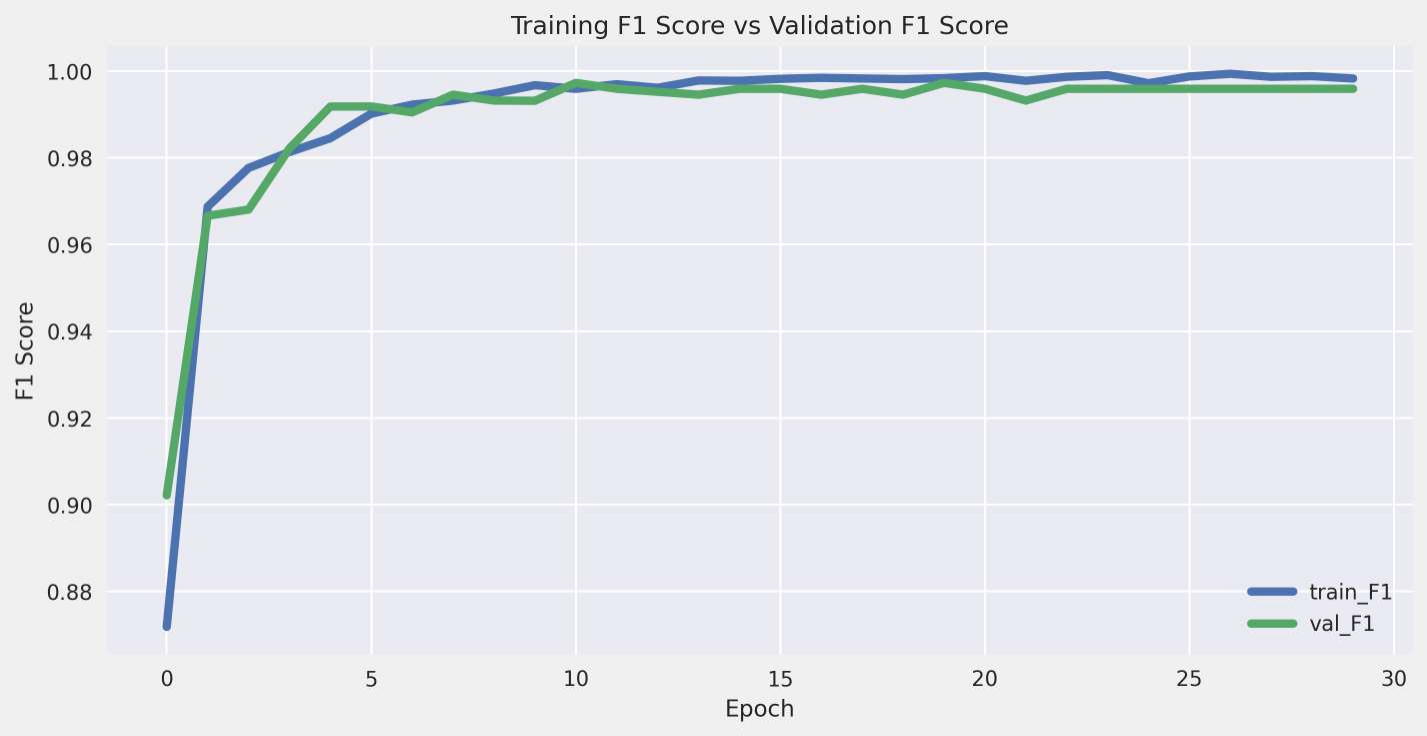}
        \caption{DenseNet 201}
    \end{subfigure}
    \hfill
    \begin{subfigure}[b]{0.45\textwidth}
        \centering
        \includegraphics[width=\textwidth]{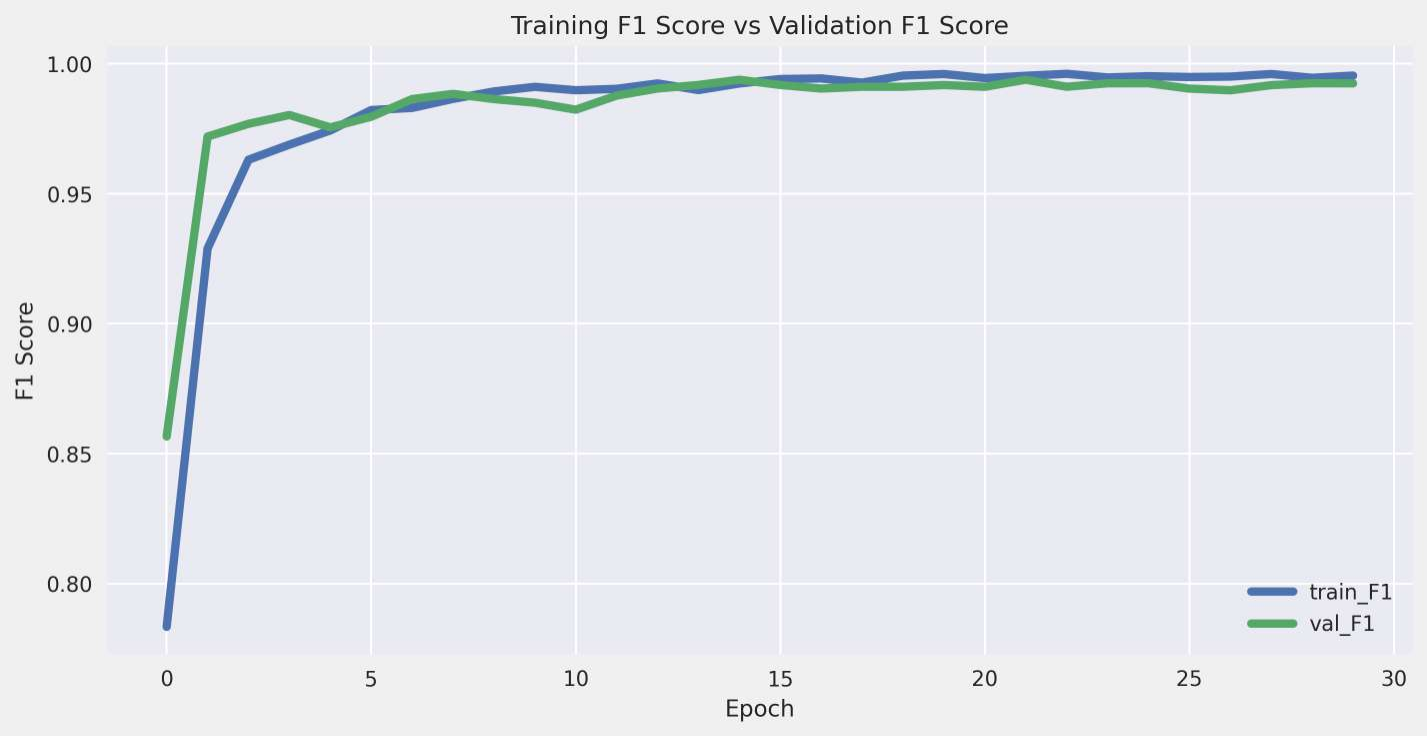}
        \caption{InceptionResNetV2}
    \end{subfigure}
    \hfill
    \begin{subfigure}[b]{0.45\textwidth}
        \centering
        \includegraphics[width=\textwidth]{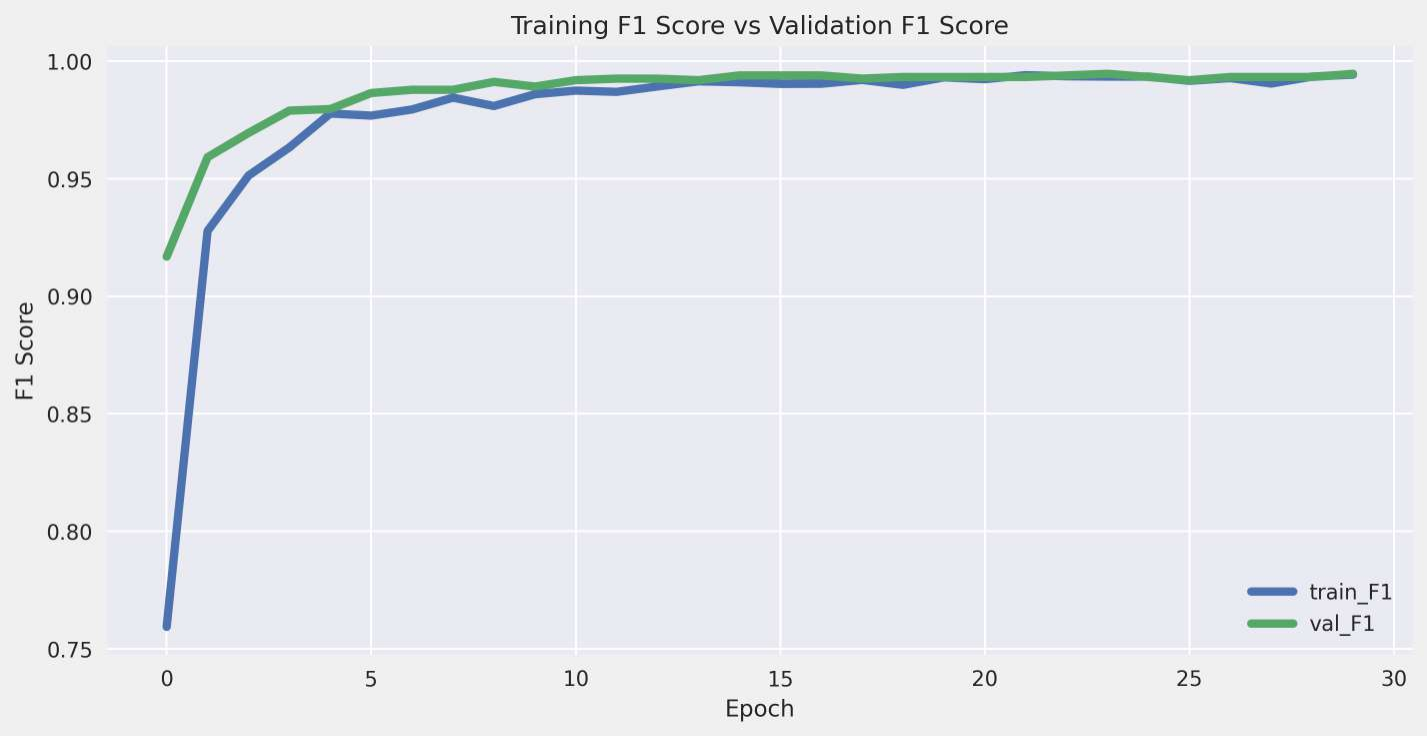}
        \caption{InceptionV3}
    \end{subfigure}
    \vfill
    \begin{subfigure}[b]{0.45\textwidth}
        \centering
        \includegraphics[width=\textwidth]{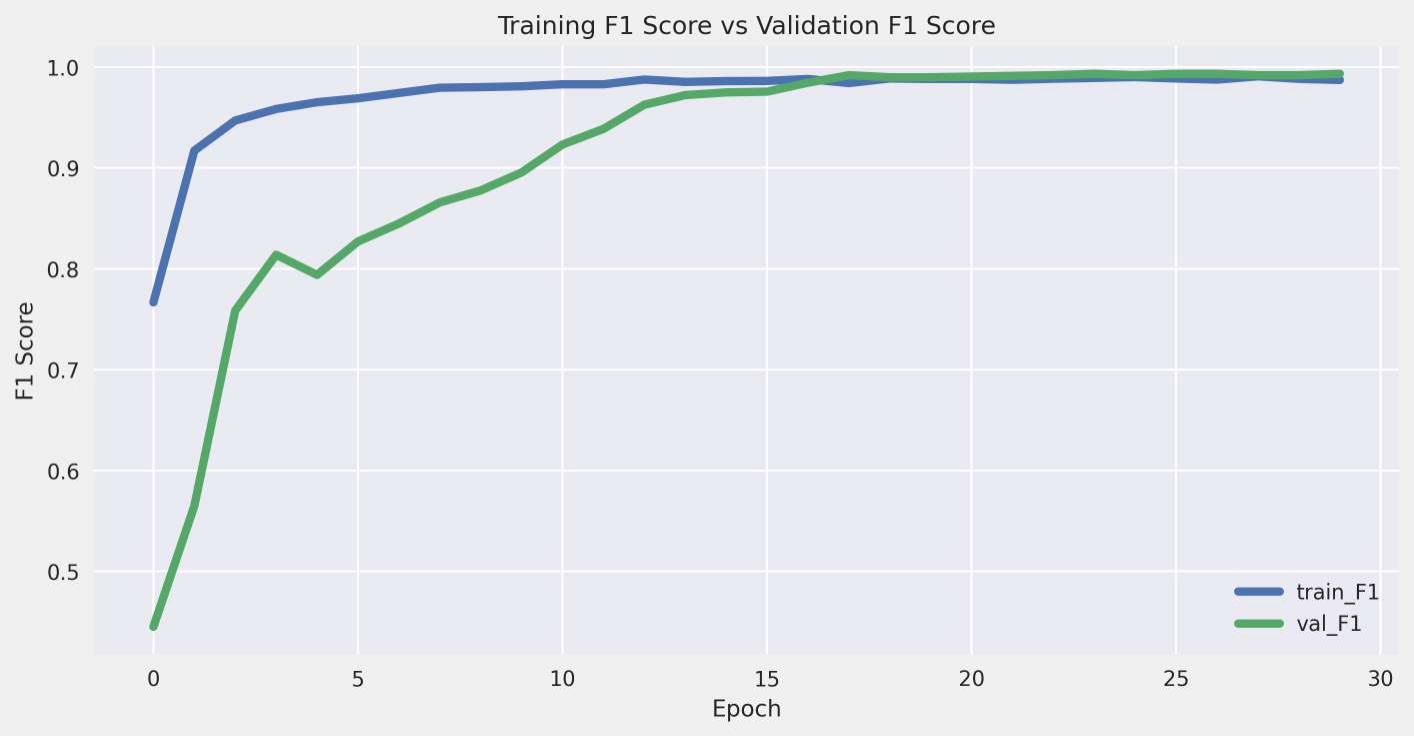}
        \caption{MobileNetV2}
    \end{subfigure}
    \hfill
    \begin{subfigure}[b]{0.45\textwidth}
        \centering
        \includegraphics[width=\textwidth]{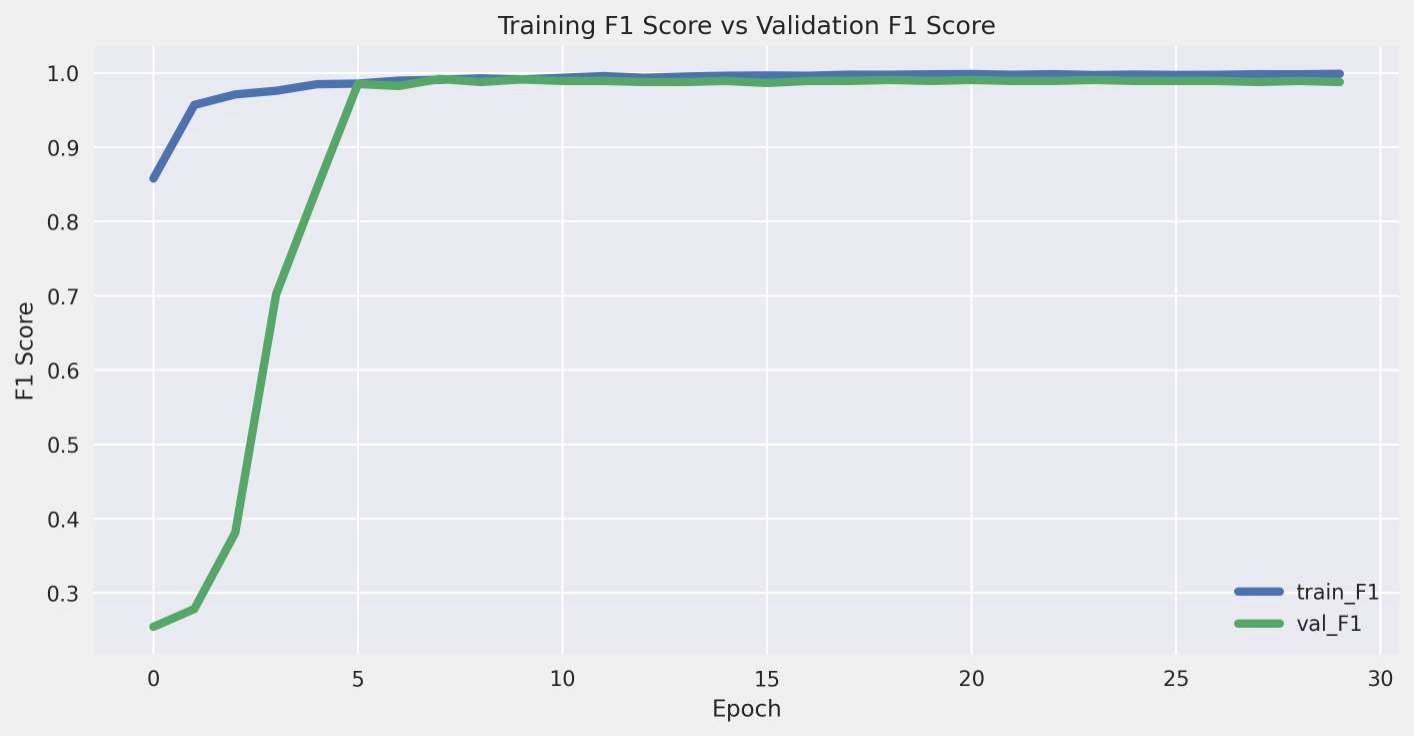}
        \caption{ResNet101}
    \end{subfigure}
    \vfill
    \begin{subfigure}[b]{0.45\textwidth}
        \centering
        \includegraphics[width=\textwidth]{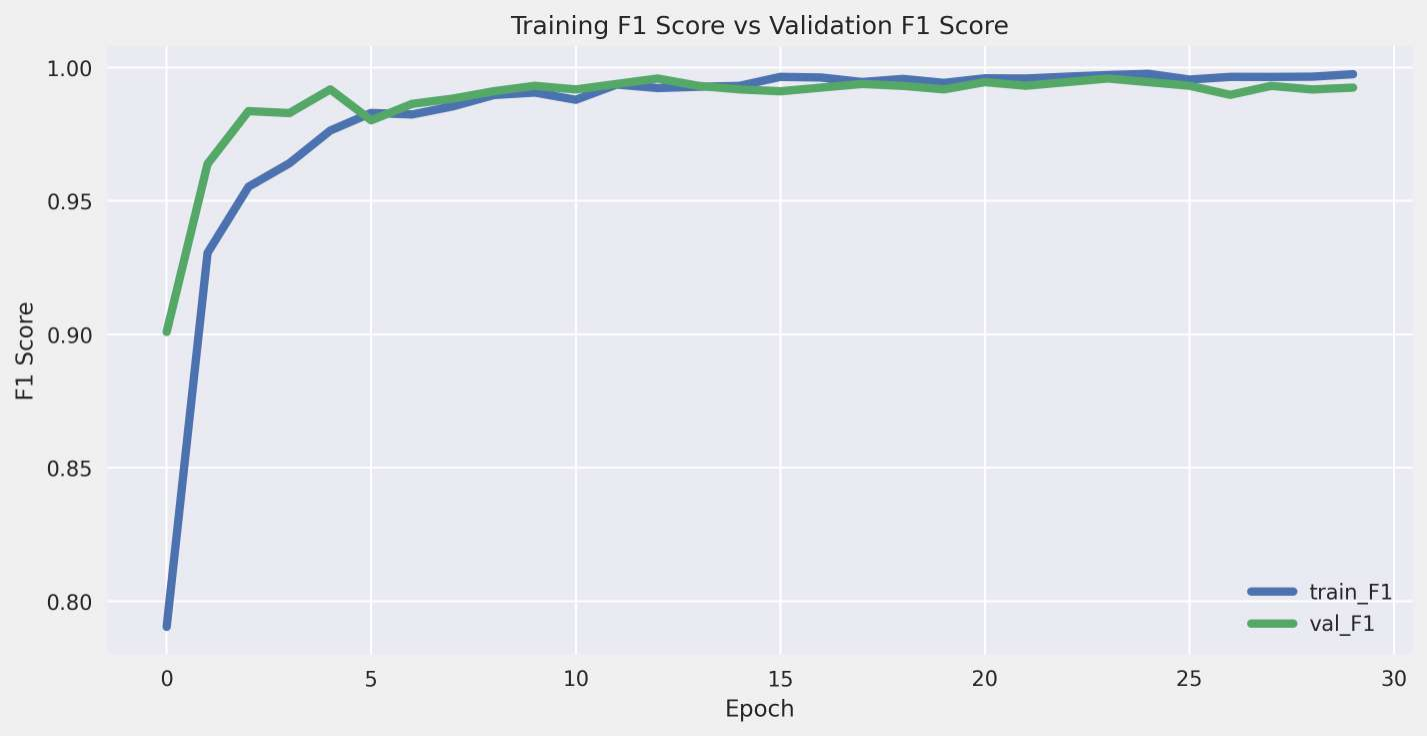}
        \caption{ResNet152 V2}
    \end{subfigure}
    \hfill
    \begin{subfigure}[b]{0.45\textwidth}
        \centering
        \includegraphics[width=\textwidth]{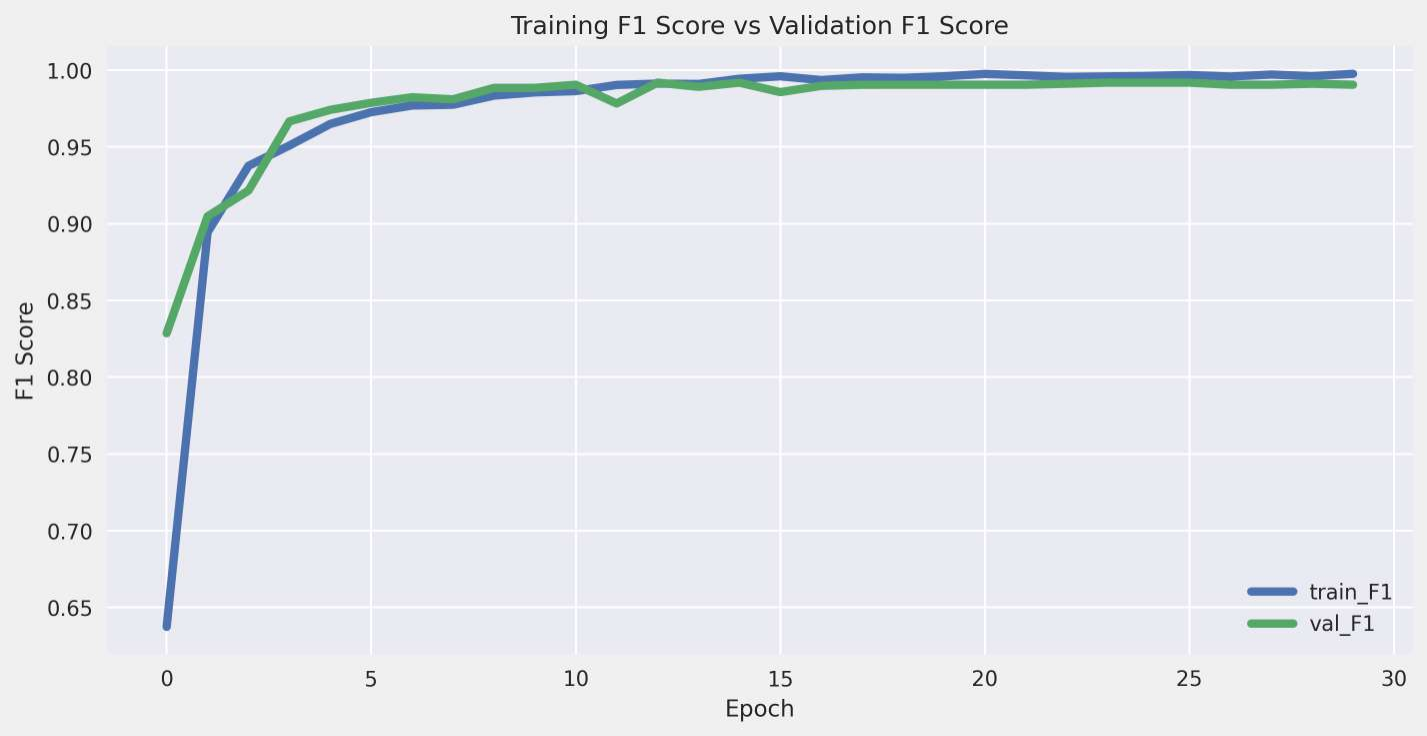}
        \caption{VGG19}
    \end{subfigure}
    \caption{Training F1 score vs validation F1 score graph}
\end{figure}
\clearpage

\clearpage
\begin{figure}[h]
    \centering
    \begin{subfigure}[b]{0.45\textwidth}
        \centering
        \includegraphics[width=\textwidth]{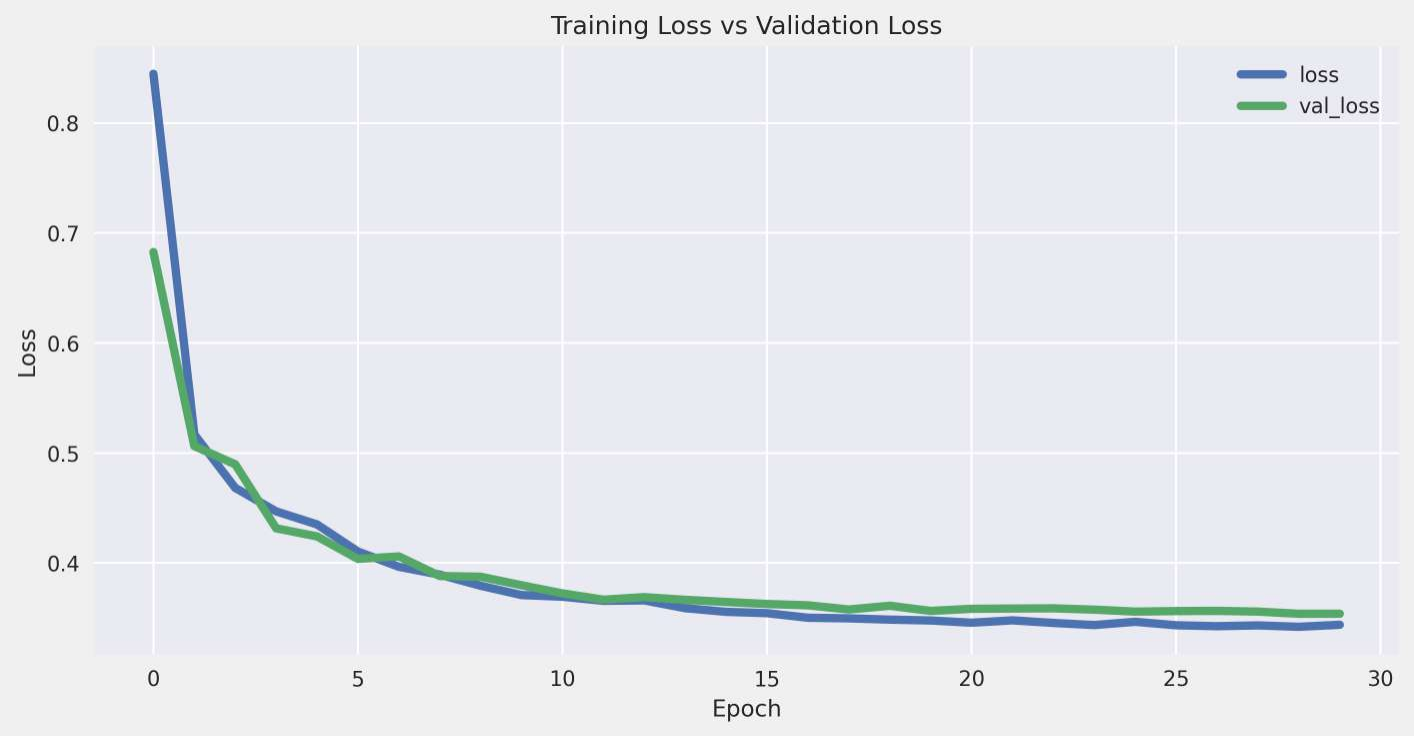}
        \caption{DenseNet 201}
    \end{subfigure}
    \hfill
    \begin{subfigure}[b]{0.45\textwidth}
        \centering
        \includegraphics[width=\textwidth]{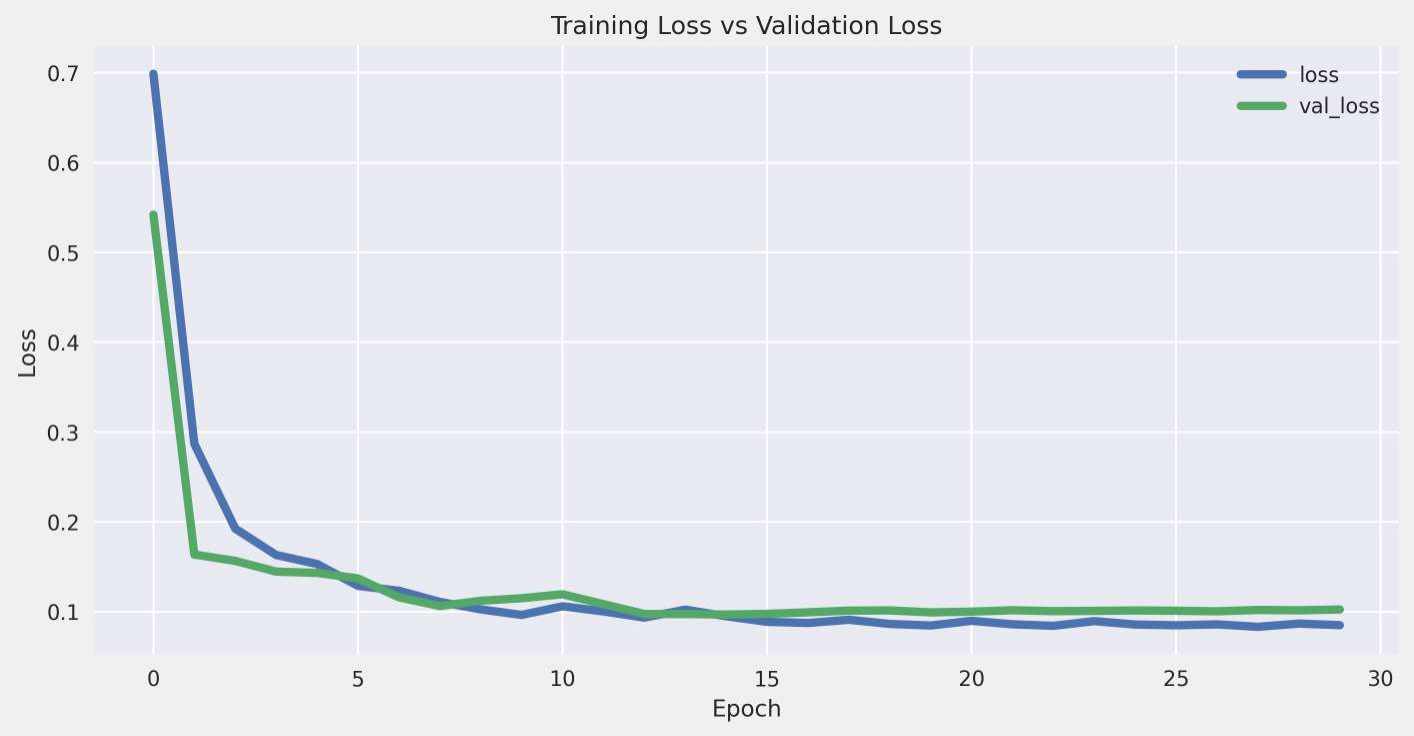}
        \caption{InceptionResNetV2}
    \end{subfigure}
    \vfill
    \begin{subfigure}[b]{0.45\textwidth}
        \centering
        \includegraphics[width=\textwidth]{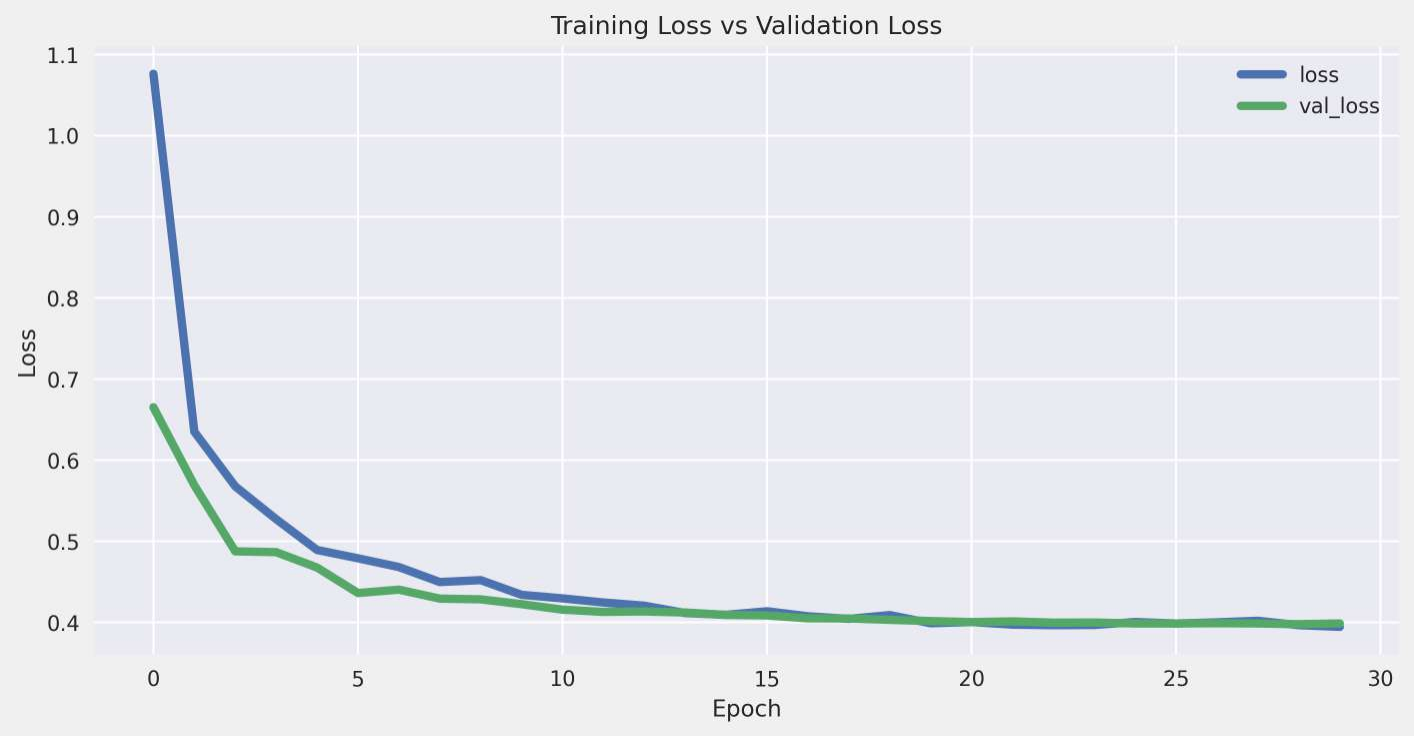}
        \caption{InceptionV3}
    \end{subfigure}
    \vfill
    \begin{subfigure}[b]{0.45\textwidth}
        \centering
        \includegraphics[width=\textwidth]{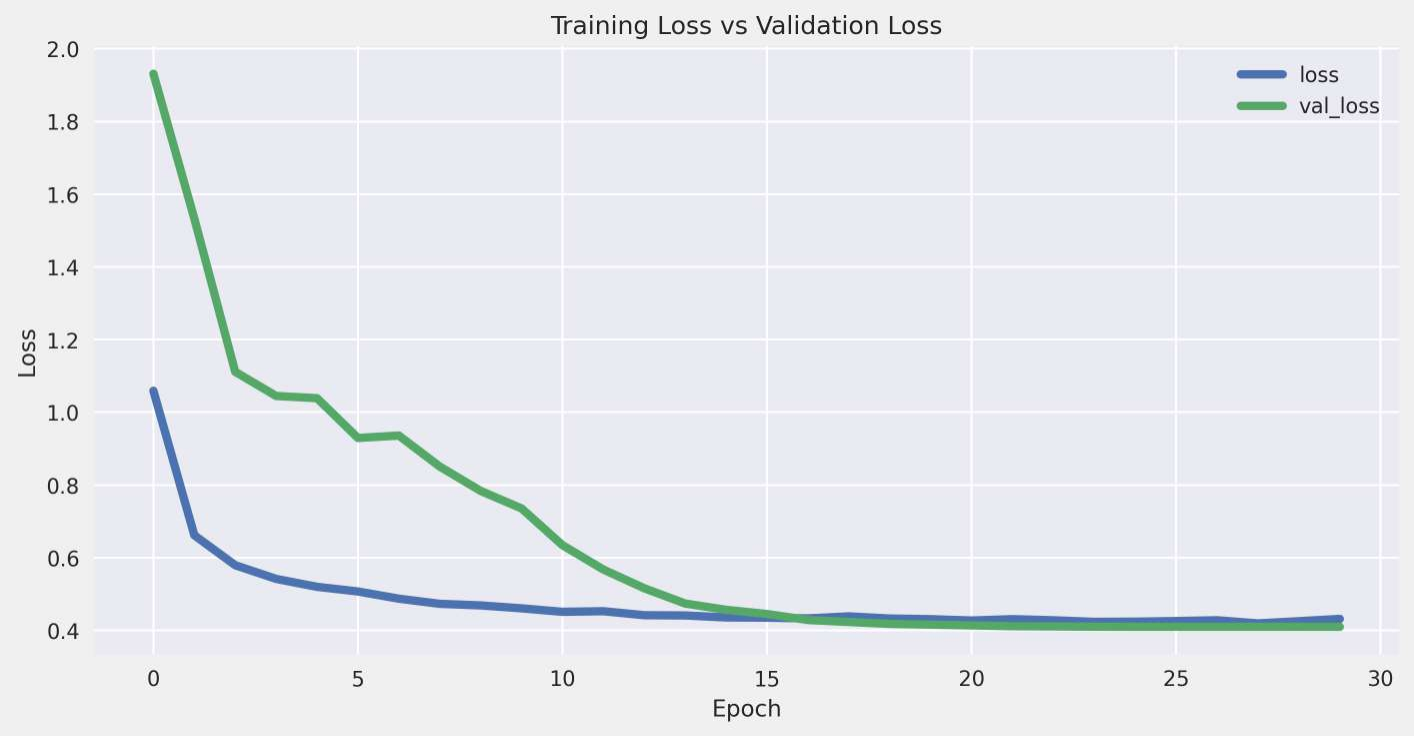}
        \caption{MobileNetV2}
    \end{subfigure}
    \hfill
    \begin{subfigure}[b]{0.45\textwidth}
        \centering
        \includegraphics[width=\textwidth]{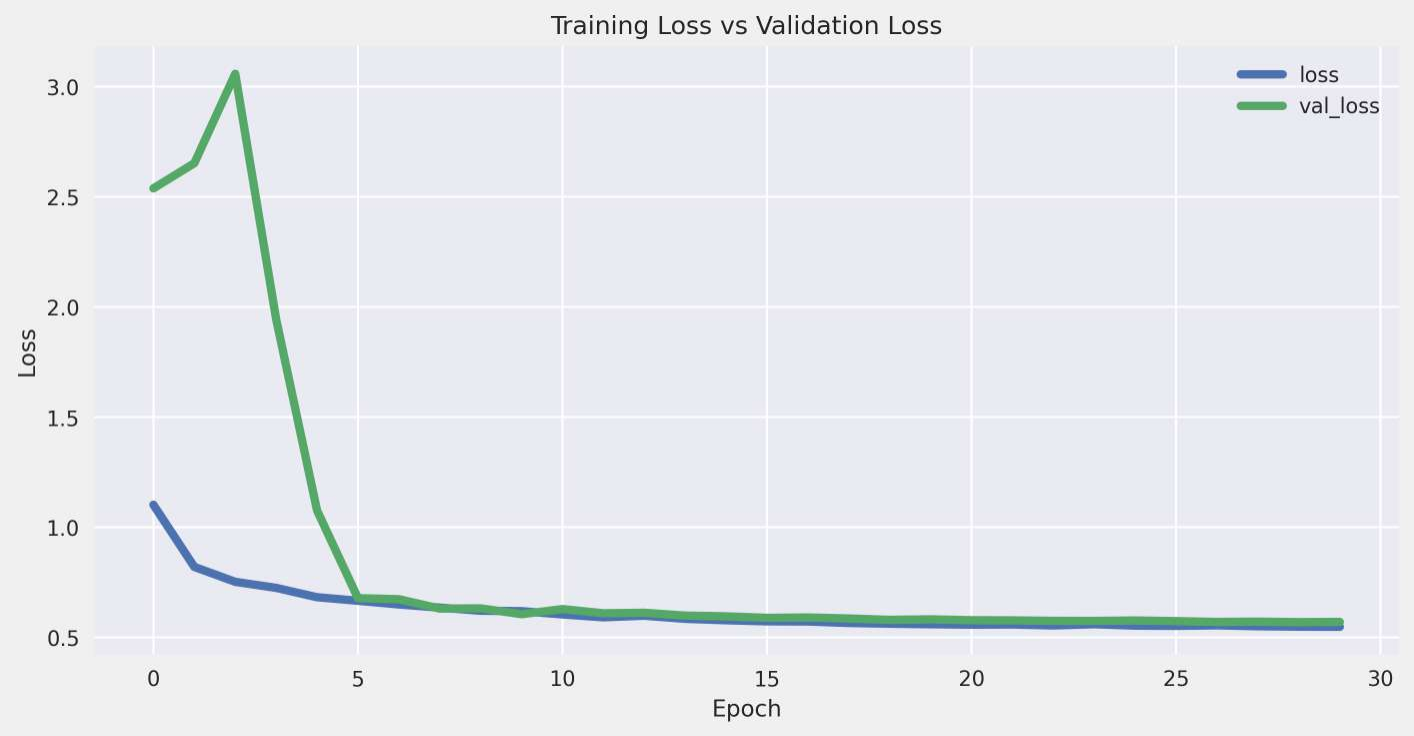}
        \caption{ResNet101}
    \end{subfigure}
    \vfill
    \begin{subfigure}[b]{0.45\textwidth}
        \centering
        \includegraphics[width=\textwidth]{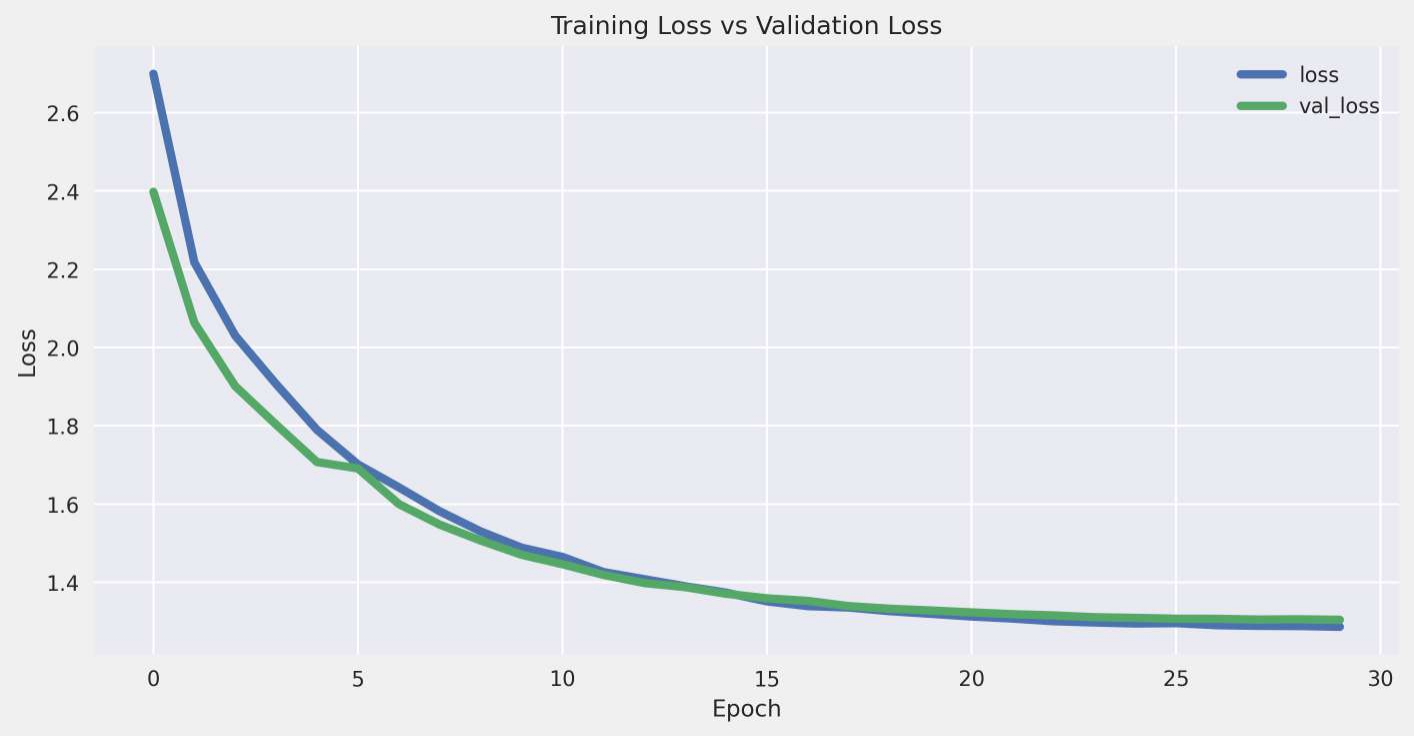}
        \caption{ResNet152 V2}
    \end{subfigure}
    \hfill
    \begin{subfigure}[b]{0.45\textwidth}
        \centering
        \includegraphics[width=\textwidth]{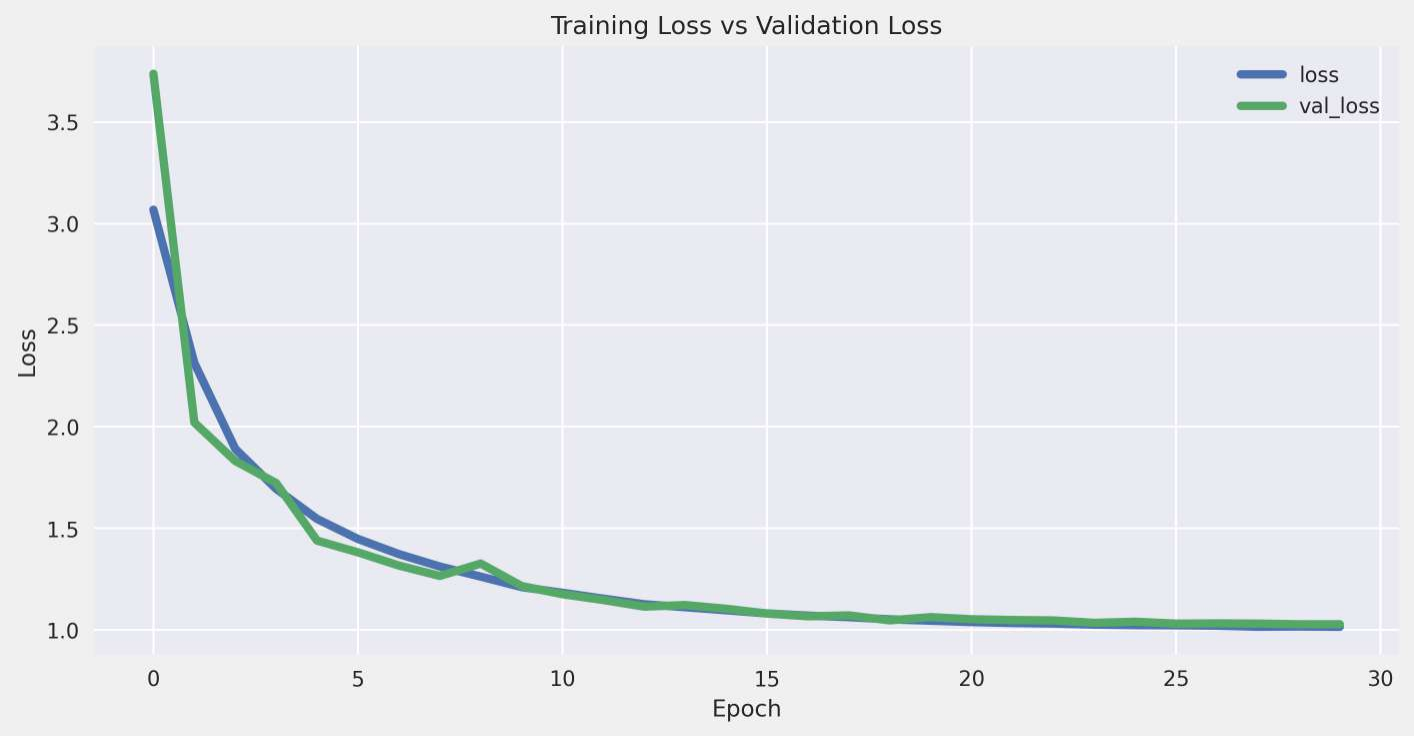}
        \caption{VGG19}
    \end{subfigure}
    \caption{Training loss vs validation loss graph}
\end{figure}
\newpage
\clearpage

\clearpage
\begin{figure*}[h]
    \centering
    \begin{subfigure}[b]{0.45\textwidth}
        \centering
        \includegraphics[width=0.75\textwidth]{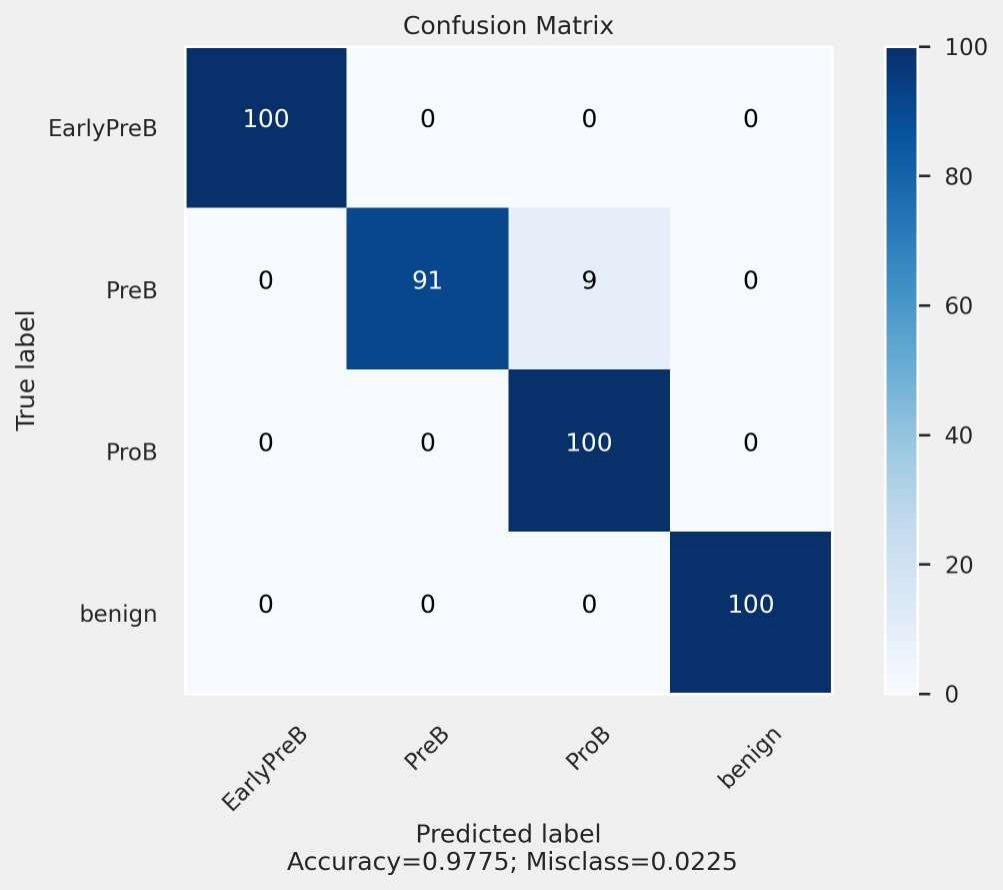}
        \caption{DenseNet 201}
    \end{subfigure}
    \begin{subfigure}[b]{0.45\textwidth}
        \centering
        \includegraphics[width=0.75\textwidth]{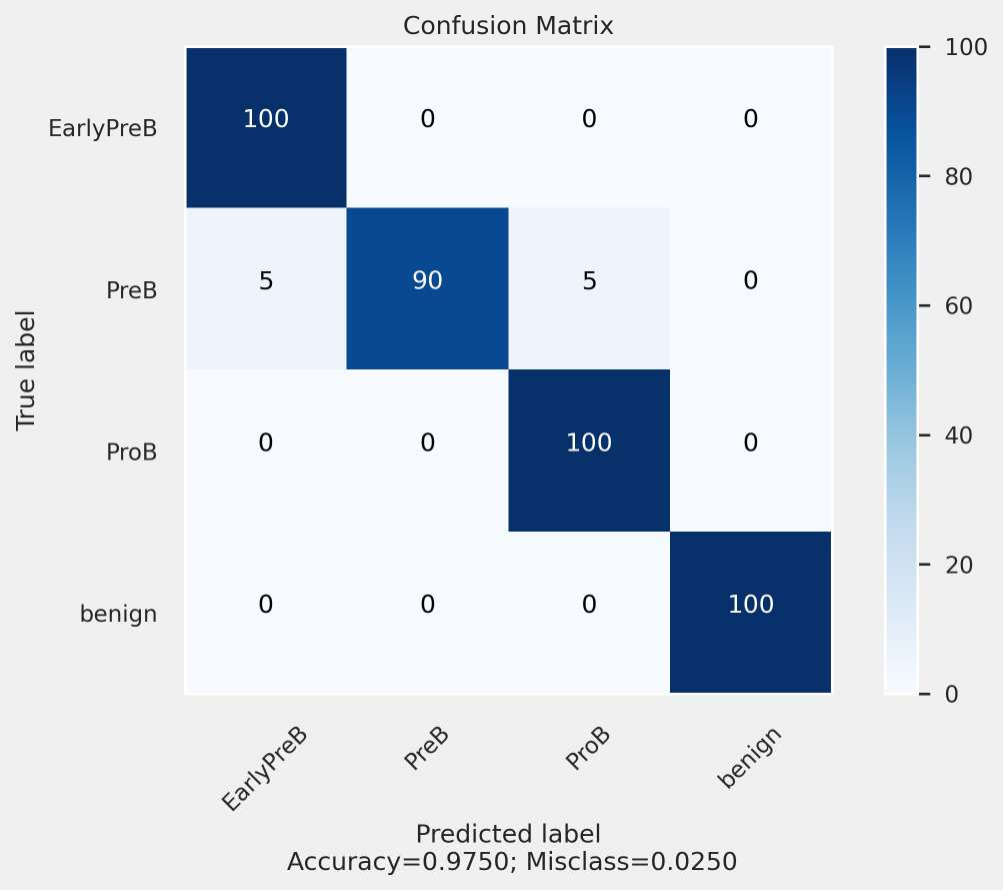}
        \caption{InceptionResNetV2}
    \end{subfigure}
    \begin{subfigure}[b]{0.45\textwidth}
        \centering
        \includegraphics[width=0.75\textwidth]{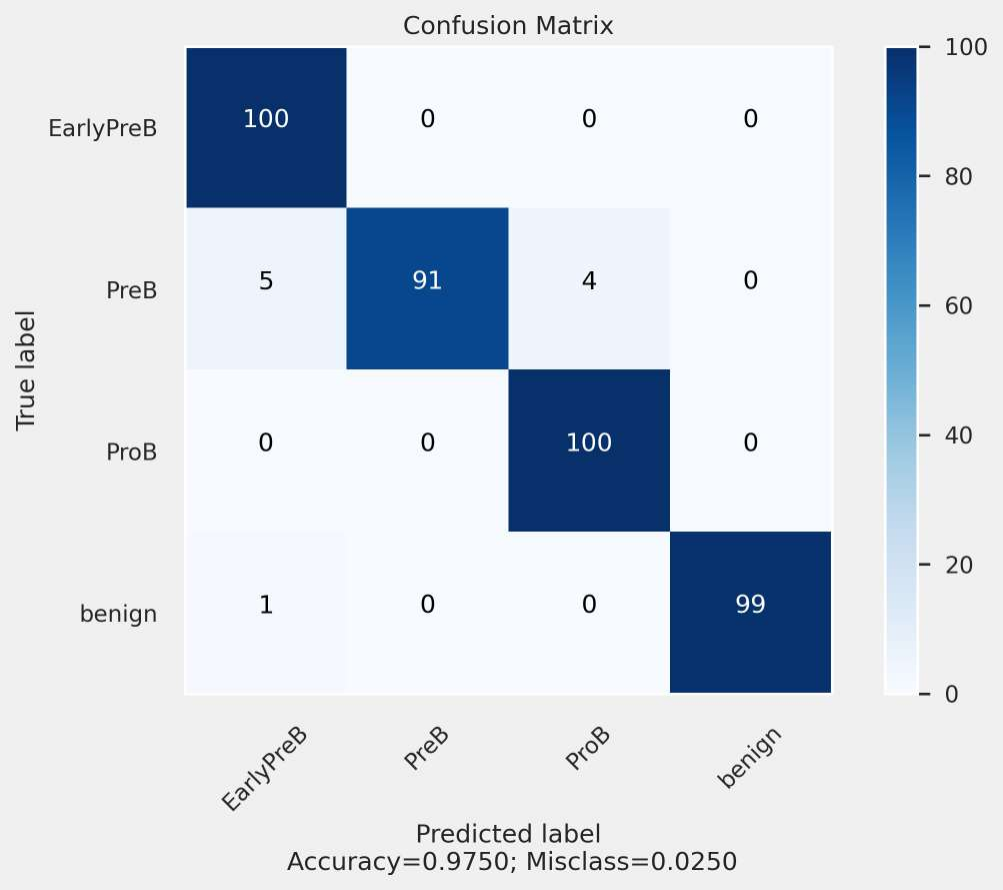}
        \caption{InceptionV3}
    \end{subfigure}
    \vfill
    \begin{subfigure}[b]{0.45\textwidth}
        \centering
        \includegraphics[width=0.75\textwidth]{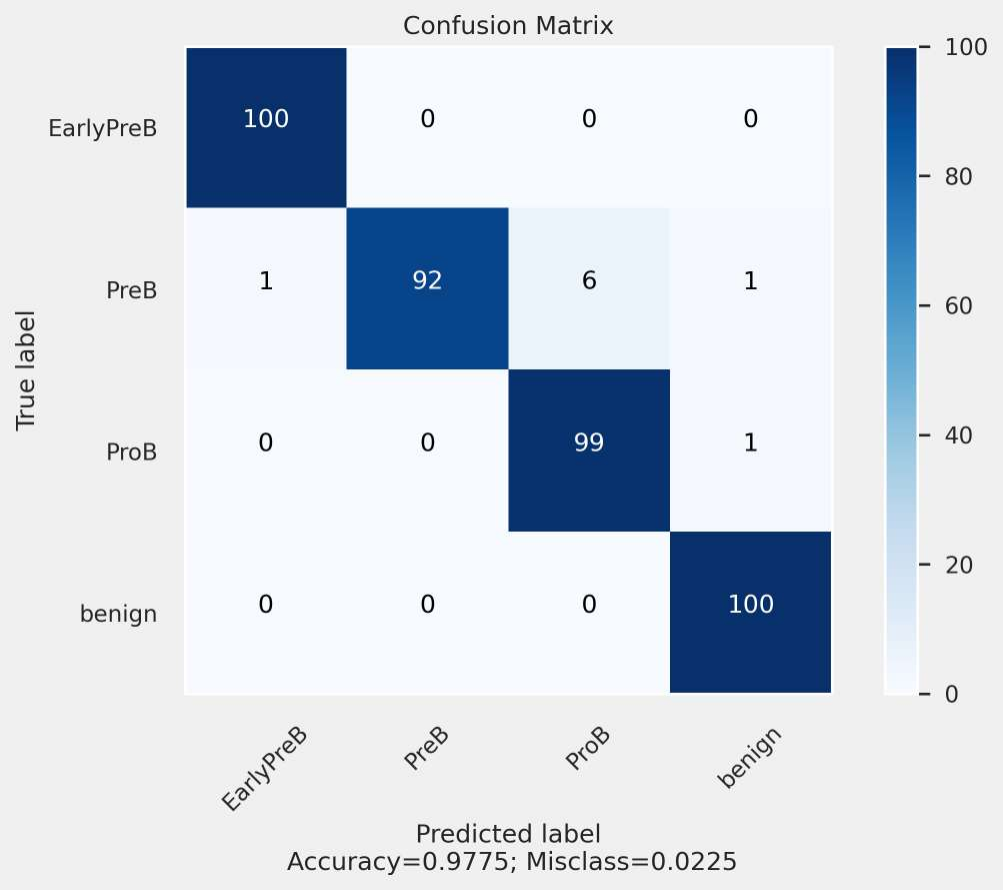}
        \caption{MobileNetV2}
    \end{subfigure}
    \begin{subfigure}[b]{0.45\textwidth}
        \centering
        \includegraphics[width=0.75\textwidth]{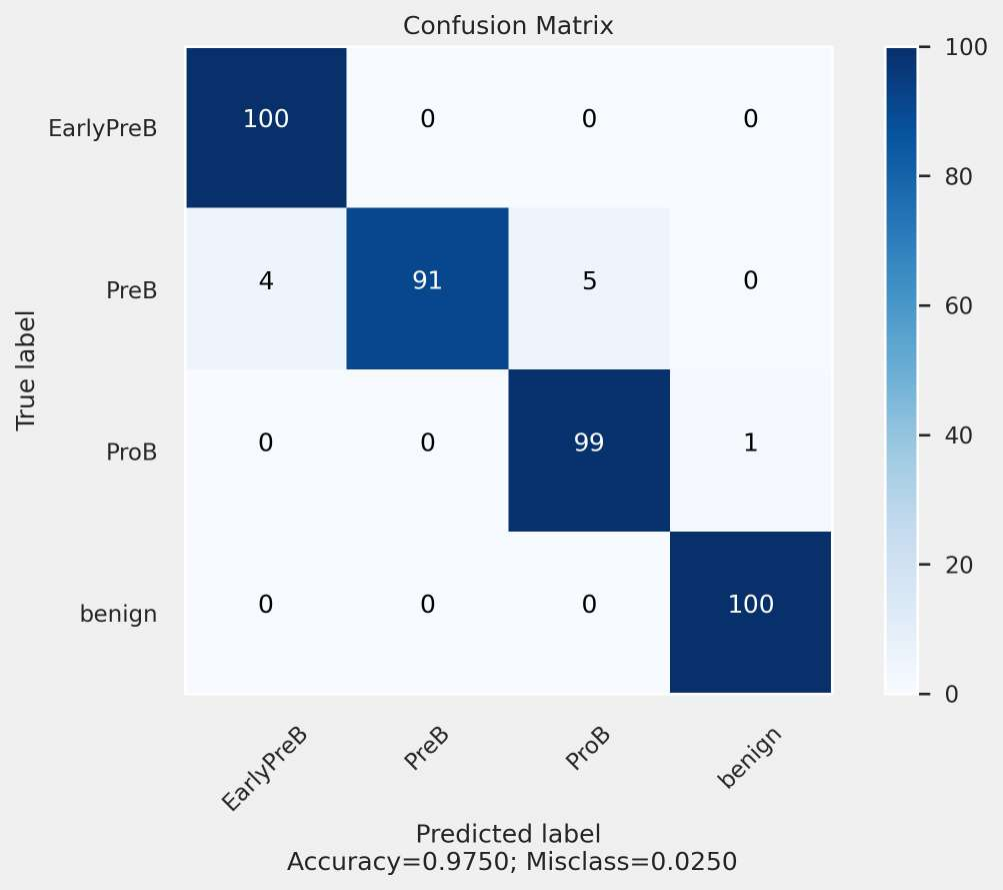}
        \caption{ResNet101}
    \end{subfigure}
    \vfill
    \begin{subfigure}[b]{0.45\textwidth}
        \centering
        \includegraphics[width=0.75\textwidth]{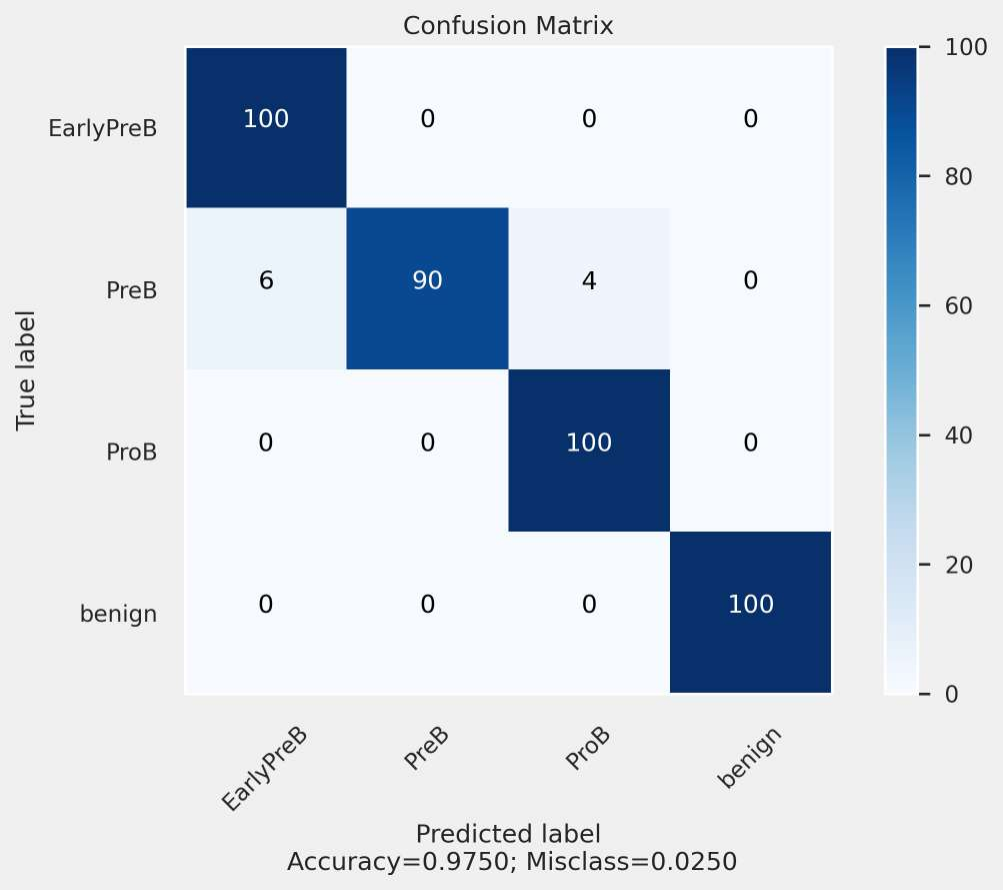}
        \caption{ResNet152 V2}
    \end{subfigure}
    \begin{subfigure}[b]{0.45\textwidth}
        \centering
        \includegraphics[width=0.75\textwidth]{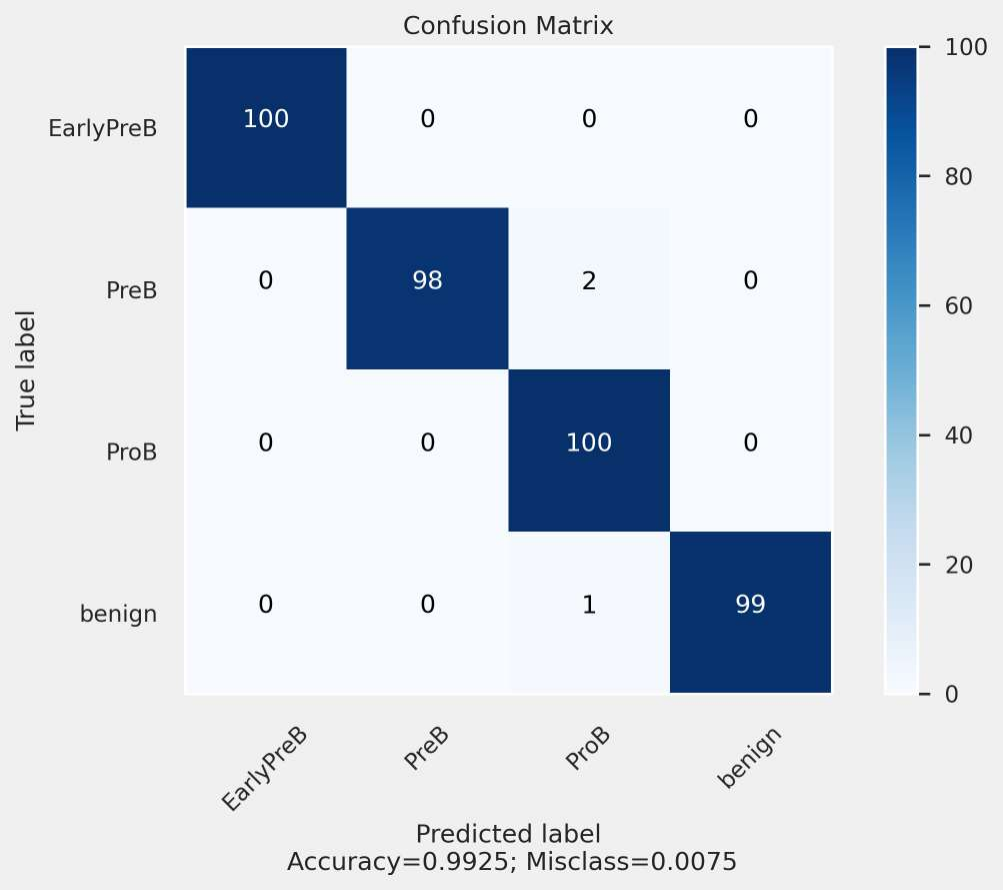}
        \caption{VGG19}
    \end{subfigure}
    \caption{Confusion matrix of test data}
\end{figure*}
\clearpage

\section{Result}
From our used models, VGG19 shows the best output among all. It exceeds others in terms of precision and accuracy. It shows quite outstanding performance for other metrics too. Other models didn't represent the best output among all, but did not lag behind too. They have also shown impressive outcomes. Performance of each model has been enlisted in in the following tables.
\begin{figure}[h]
	\centering
	\includegraphics[width=\textwidth]{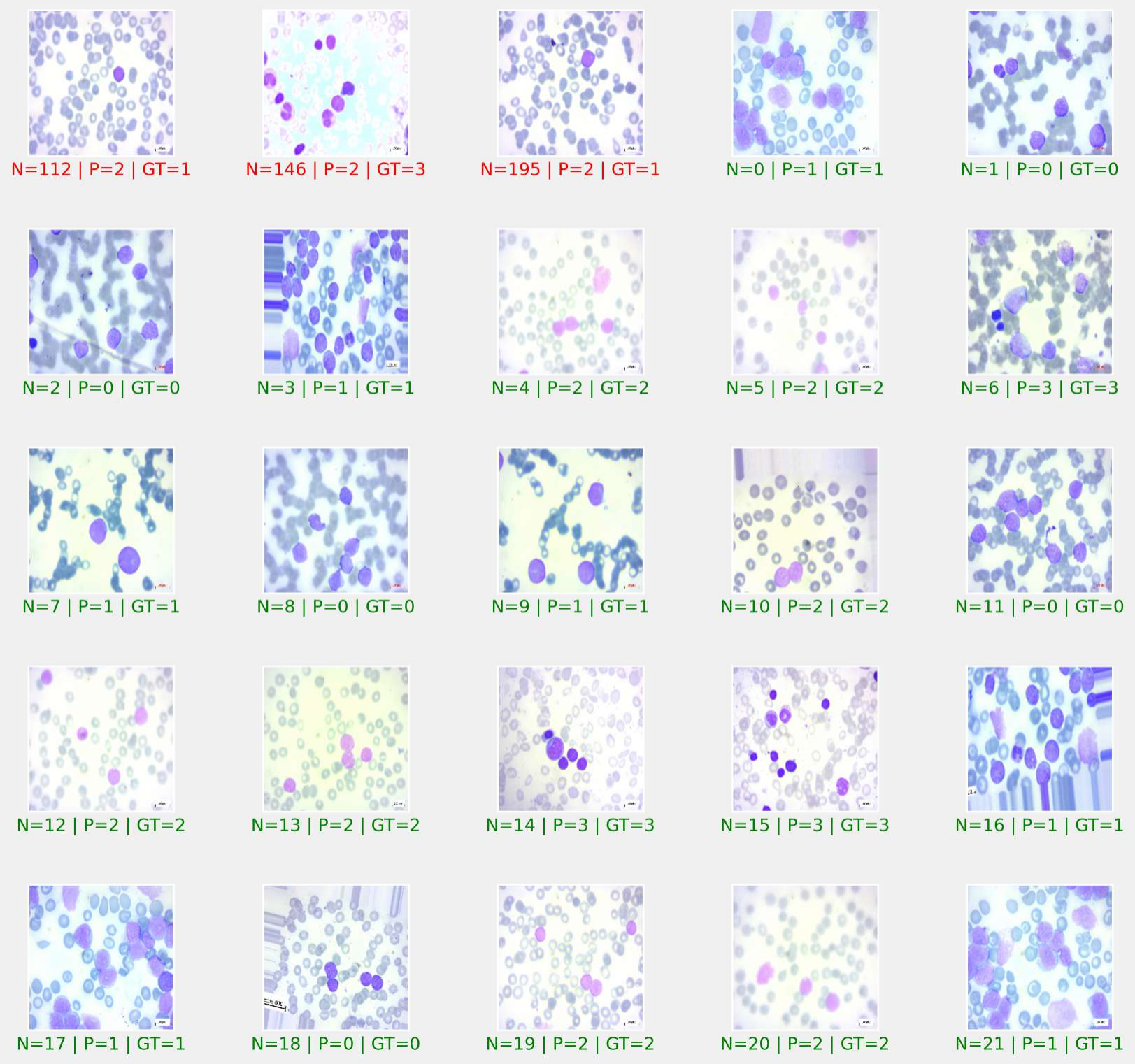}
	\caption{ Inaccuracy of VGG19 in test dataset}
\end{figure}

\noindent We discovered some instances of misclassification in our test data during our analysis, where our used models mistakenly detected a class that should not have been present in the image. On closer inspection, it was discovered that the photograph in question had poor clarity and a pronounced yellowish hue. The model's correct classification was hindered by this intrinsic visual uncertainty, which was most likely caused by image quality or lighting circumstances. Such examples highlight the significance of high-quality data and the impact of visual nuances on deep-learning model performance in medical image processing. This observation contributes to our ongoing attempts to improve model robustness in the face of fluctuating picture features and quality, which is an important component of our research focus.
\begin{table}[H]
	\centering
	\caption{Result on test dataset with custom architecture}
	\begin{tabular}{c c c c c c c}
		\hline
		\textbf{\begin{tabular}[c]{@{}c@{}}Model name\end{tabular}} & \textbf{Accuracy} & \textbf{Loss}               & \textbf{AUC}    & \textbf{Recall} & \textbf{Precision} & \textbf{F1 score} \\ \hline
		\textbf{InceptionResNetV2}         & {0.9750}   & 	0.2446 & 	0.9900          & 0.9750          & 0.9750             & 0.9760            \\ \hline
		\textbf{ResNet101}                 & {0.9750}   & 0.6832                      & 0.9925          & {0.9750} & 0.9750             & 0.9760            \\ \hline
		\textbf{ResNet152V2}               & {0.9750}   &{1.4414}             & {0.9898} & {0.9725} & \textbf{0.9749}     & \textbf{0.9747}   \\ \hline
		\textbf{MobileNetV2}               & 0.9750            & 1.2851                      & 0.9912          & 0.9748          & 0.9750             & 0.9758            \\ \hline
		\textbf{DenseNet201}               & 0.9775            & 0.4576                      & 0.9925          & 0.9775          & 0.9775              & 0.9784           \\ \hline
		\textbf{InceptionV3}               & 0.9750            & 0.5889                      & 0.9925          & 0.9750          & 9750              & 0.9762            \\ \hline
		
		\textbf{VGG19}                     & \textbf{0.9925}            & 0.9865                      & 0.9982          & 0.9900          & 0.9925             & 0.9915            \\ \hline
	\end{tabular}
\end{table}

\begin{table}[H]
	\centering
	\caption{Result on test dataset without custom architecture}
	\begin{tabular}{c c c c c c c}
		\hline
		\textbf{\begin{tabular}[c]{@{}c@{}}Model name\end{tabular}} & \textbf{Accuracy} & \textbf{Loss}               & \textbf{AUC}    & \textbf{Recall} & \textbf{Precision} & \textbf{F1 score} \\ \hline
		\textbf{InceptionResNetV2}         & \textbf{0.9800}   & 	0.1152 & 	0.9939          & 0.9800          & 0.9800             & 0.9800            \\ \hline
		\textbf{ResNet101}                 & {0.9775}   & 0.3222                      & 0.9873          & \textbf{0.9775} & 0.9775             & 0.9784            \\ \hline
		\textbf{ResNet152V2}               & {0.9775}   & {0.1618}             & {0.9891} & {0.9775} & {0.9775}     & {0.9784}   \\ \hline
		\textbf{MobileNetV2}               & 0.9725            & 0.1573                      & 0.9913          & 0.9725          & 0.9725            & 0.9736            \\ \hline
		\textbf{DenseNet201}               & 0.9750            & 0.1384                      & 0.9930          & 0.9750          & 0.9750              & 0.9760            \\ \hline
		\textbf{InceptionV3}               & 0.9650            & 0.2321                      & 0.9872          & 0.9650          & 0.9650              & 0.9663            \\ \hline
		
		\textbf{VGG19}                     & \textbf{0.9800}           & 0.0917                     & 0.9947          & 0.9800           & 0.9800              & 0.9784            \\ \hline
	\end{tabular}
\end{table}

\begin{landscape}
    \begin{table}[h]
        \centering
        \caption{Result (Performance of different models during training with custom architecture)}
        \vspace{1em}
        \small 
        \begin{tabular}{|c|c|cc|c|cc|cc|cc|}
            \hline
            \multirow{2}{*}{\textbf{\begin{tabular}[c]{@{}c@{}}Model\\ name\end{tabular}}} & \multirow{2}{*}{\textbf{\begin{tabular}[c]{@{}c@{}}Batch\\ Size\end{tabular}}} & \multicolumn{2}{c|}{\textbf{\begin{tabular}[c]{@{}c@{}}Custom model accuracy\\ (after 30 epochs)\end{tabular}}} & \multirow{2}{*}{\textbf{Accuracy}} & \multicolumn{2}{c|}{\textbf{Recall}} & \multicolumn{2}{c|}{\textbf{Precision}} & \multicolumn{2}{c|}{\textbf{F1 score}} \\ \cline{3-4} \cline{6-11}
            & & \multicolumn{1}{c|}{\textbf{Training}} & \textbf{Validation} & & \multicolumn{1}{c|}{\textbf{Training}} & \textbf{Validation} & \multicolumn{1}{c|}{\textbf{Training}} & \textbf{Validation} & \multicolumn{1}{c|}{\textbf{Training}} & \textbf{Validation} \\ \hline
            InceptionResNetV2 & 32 & \multicolumn{1}{c|}{0.9954} & 0.9903 & {0.9750} & \multicolumn{1}{c|}{0.9952} & 0.9903 & \multicolumn{1}{c|}{0.9957} & 0.9944 & \multicolumn{1}{c|}{0.9955} & 0.9925 \\ \hline
            ResNet101 & 32 & \multicolumn{1}{c|}{0.9988} & 0.9889 & {0.9750} & \multicolumn{1}{c|}{0.9986} & 0.9889 & \multicolumn{1}{c|}{{0.9989}} & 0.9889 & \multicolumn{1}{c|}{0.9988} & 0.9878 \\ \hline
            ResNet152V2 & 32 & \multicolumn{1}{c|}{{0.9977}} & \textbf{0.9917} & {0.9750} & \multicolumn{1}{c|}{{0.9974}} &{0.9917} & \multicolumn{1}{c|}{{0.9977}} & 0.9930 & \multicolumn{1}{c|}{{0.9975}} & 0.9925 \\ \hline
            MobileNetV2 & 32 & \multicolumn{1}{c|}{0.9974} & 0.9949 & 0.9786 & \multicolumn{1}{c|}{0.9970} & 0.9949 & \multicolumn{1}{c|}{0.9976} & 0.9949 & \multicolumn{1}{c|}{0.9973} & 0.9951 \\ \hline
            DenseNet201 & 32 & \multicolumn{1}{c|}{0.9983} & {0.9958} & 0.9775 & \multicolumn{1}{c|}{0.9981} & 0.9958 & \multicolumn{1}{c|}{0.9985} & {0.9958} & \multicolumn{1}{c|}{0.9983} & {0.9959} \\ \hline
            InceptionV3 & 32 & \multicolumn{1}{c|}{0.9975} & 0.9972 & 0.9750 & \multicolumn{1}{c|}{0.9975} & 0.9983 & \multicolumn{1}{c|}{0.9979} & 0.9983 & \multicolumn{1}{c|}{0.9977} & 0.9984 \\ \hline
            
            VGG19 & 32 & \multicolumn{1}{c|}{0.9958} & 0.9875 & 0.9925 & \multicolumn{1}{c|}{0.9955} & 0.9861 & \multicolumn{1}{c|}{0.9964} & 0.9875 & \multicolumn{1}{c|}{0.9960} & 0.9871 \\ \hline
        \end{tabular}
    \end{table}

    \vspace{1em}
    
      \begin{table}[h]
        \centering
        \caption{Result (Performance of different models during training without custom architecture)}
        \vspace{1em}
        \small 
        \begin{tabular}{|c|c|cc|c|cc|cc|cc|}
            \hline
            \multirow{2}{*}{\textbf{\begin{tabular}[c]{@{}c@{}}Model\\ name\end{tabular}}} & \multirow{2}{*}{\textbf{\begin{tabular}[c]{@{}c@{}}Batch\\ Size\end{tabular}}} & \multicolumn{2}{c|}{\textbf{\begin{tabular}[c]{@{}c@{}}Custom model accuracy\\ (after 30 epochs)\end{tabular}}} & \multirow{2}{*}{\textbf{Accuracy}} & \multicolumn{2}{c|}{\textbf{Recall}} & \multicolumn{2}{c|}{\textbf{Precision}} & \multicolumn{2}{c|}{\textbf{F1 score}} \\ \cline{3-4} \cline{6-11}
            & & \multicolumn{1}{c|}{\textbf{Training}} & \textbf{Validation} & & \multicolumn{1}{c|}{\textbf{Training}} & \textbf{Validation} & \multicolumn{1}{c|}{\textbf{Training}} & \textbf{Validation} & \multicolumn{1}{c|}{\textbf{Training}} & \textbf{Validation} \\ \hline
            InceptionResNetV2 & 32 & \multicolumn{1}{c|}{0.9997} & 0.9903 & {0.9800} & \multicolumn{1}{c|}{0.9997} & 0.9903 & \multicolumn{1}{c|}{0.9997} & 0.9903 & \multicolumn{1}{c|}{0.9997} & 0.9903 \\ \hline
            ResNet101 & 32 & \multicolumn{1}{c|}{1.000} & 0.9972 & {0.9775} & \multicolumn{1}{c|}{1.000} & 0.9972 & \multicolumn{1}{c|}{{1.000}} & 0.9972 & \multicolumn{1}{c|}{0.9993} & 0.9973 \\ \hline
            ResNet152V2 & 32 & \multicolumn{1}{c|}{{0.9994}} & {0.9944} & {0.9775} & \multicolumn{1}{c|}{{0.9994}} & {0.9944} & \multicolumn{1}{c|}{{0.9994}} & 0.9944 & \multicolumn{1}{c|}{0.9994} & 0.9946 \\ \hline
            MobileNetV2 & 32 & \multicolumn{1}{c|}{0.9991} & 0.9958 & 0.9725 & \multicolumn{1}{c|}{0.9991} & 0.9958 & \multicolumn{1}{c|}{0.9991} & 0.9958 & \multicolumn{1}{c|}{0.9989} & 0.9959 \\ \hline
            DenseNet201 & 32 & \multicolumn{1}{c|}{1.0000} & {0.9958} & 0.9750 & \multicolumn{1}{c|}{1.0000} & 0.9958 & \multicolumn{1}{c|}{0.9992} & {0.9958} & \multicolumn{1}{c|}{1.0000} & {0.9959} \\ \hline
            InceptionV3 & 32 & \multicolumn{1}{c|}{0.9992} & 0.9931 & 0.9650 & \multicolumn{1}{c|}{0.9991} & 0.9931 & \multicolumn{1}{c|}{0.9992} & 0.9944 & \multicolumn{1}{c|}{0.9992} & 0.9939 \\ \hline
            
            VGG19 & 32 & \multicolumn{1}{c|}{0.9997} & 0.9931 & 0.9800 & \multicolumn{1}{c|}{0.9997} & 0.9931 & \multicolumn{1}{c|}{0.9997} & 0.9931 & \multicolumn{1}{c|}{0.9997} & 0.9932 \\ \hline
        \end{tabular}
    \end{table}
\end{landscape}

\section{Future Scopes}
The integration of machine learning into medical imaging research has historically encountered challenges, particularly the limited size and diversity of datasets compared to other fields \cite{litjens2017survey}. To mitigate this, our study employed data augmentation techniques, such as image rotation, to enhance dataset variability and model generalization. Looking forward, the adoption of generative models represents a transformative opportunity in this domain. By synthesizing high-quality, realistic medical images, generative adversarial networks (GANs) and diffusion models can not only expand dataset sizes but also enhance the robustness and accuracy of diagnostic algorithms \cite{fridadar2018gan, kazerouni2022diffusion}. This approach could lead to significant advancements in the reliability of automated detection and treatment planning systems. Furthermore, the implementation of cloud-based ALL detection systems emerges as a critical step in modernizing diagnostic workflows. Cloud platforms offer unparalleled scalability, accessibility, and compliance with data security standards (such as HIPAA and GDPR), enabling remote diagnostics, real-time analytics, and collaborative healthcare decision-making \cite{razzak2020bigdata, topol2019highperformance}. Building on this, the proposed model can be translated into a secure, interactive web application that allows clinicians and pathologists to upload bone marrow smear images, receive instant AI-driven diagnoses, visualize attention-based interpretability, and generate comprehensive reports. Such a system, powered by frameworks like React.js on the frontend and Flask or FastAPI with PyTorch or TensorFlow on the backend, can be efficiently deployed on cloud infrastructure (e.g., AWS, GCP) with GPU acceleration \cite{krishnan2020scalable}. Additionally, mobile applications can further democratize access to AI diagnostics, enabling remote image capture and submission for cloud-based inference or even offline analysis using lightweight, on-device model variants (e.g., TensorFlow Lite or ONNX) \cite{howard2019mobilenetv3}. For resource-constrained environments, embedding the model into portable diagnostic hardware—such as AI-integrated microscopes using platforms like NVIDIA Jetson Nano or Raspberry Pi with Coral TPU—can facilitate offline, real-time ALL detection at the point of care \cite{ching2018obstacles}. Across all implementations, model explainability, robustness to diverse imaging conditions, and regulatory compliance will be essential to foster trust and ensure clinical efficacy \cite{holzinger2021explainable}. Collectively, these advancements point toward a future where AI-powered diagnostic systems are not only highly accurate and interpretable but also widely accessible, cost-effective, and seamlessly integrated into global healthcare infrastructures.

\begin{figure}[h]
	\centering
	\includegraphics[width=0.7\textwidth]{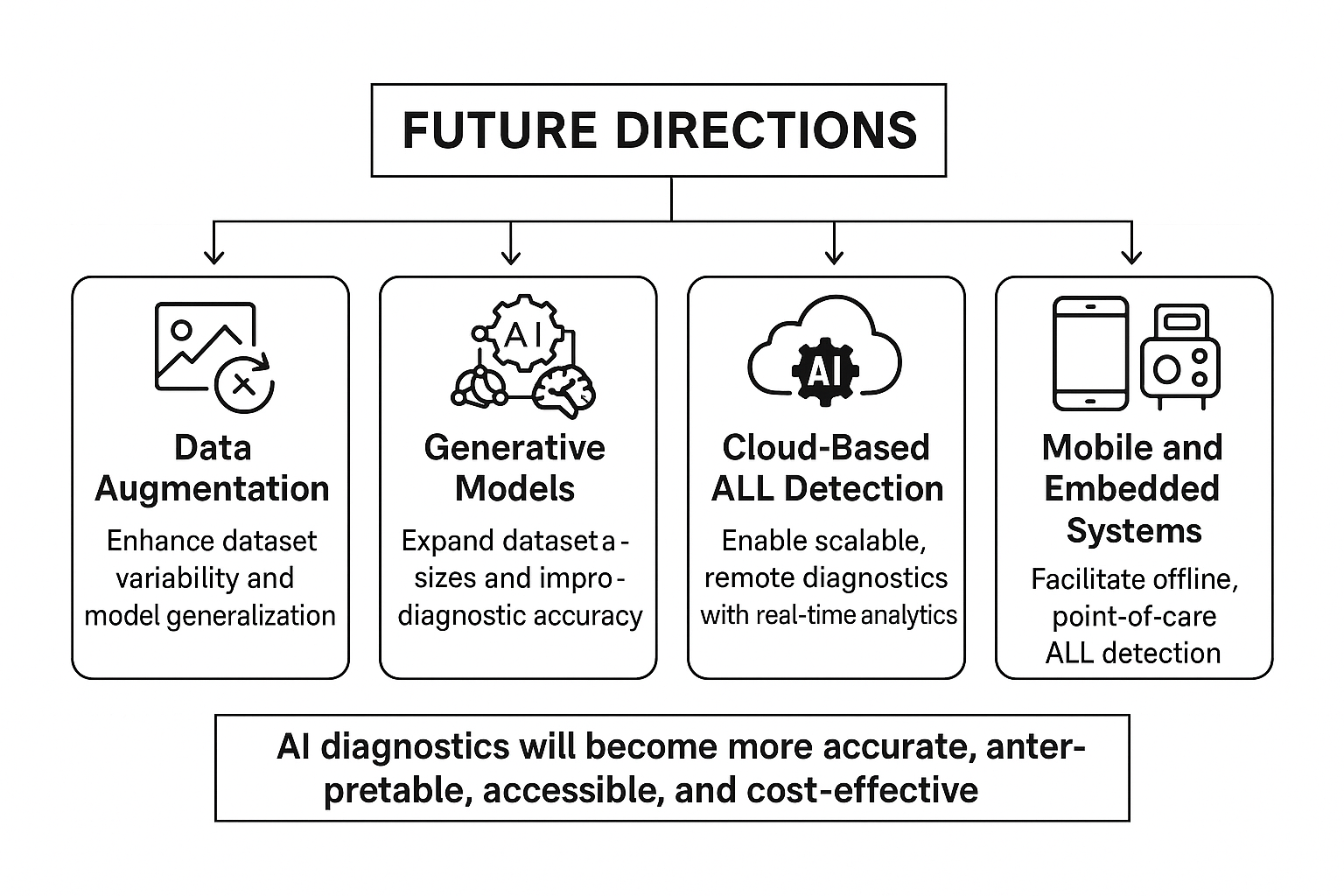}
	\caption{Future Directions for AI-Powered Acute Lymphoblastic Leukemia Diagnosis}
\end{figure}

\section{Conclusion}
In conclusion, this research article has comprehensively addressed the critical challenge of Acute Lymphoblastic Leukemia (ALL) detection from bone marrow smear images, a field of paramount importance with far-reaching implications for patient diagnosis and treatment. By undertaking a systematic investigation combining literature review, model development, and performance evaluation, this study has demonstrated the transformative potential of deep learning technologies in advancing hematological diagnostics.

The investigation began with an extensive review of existing methodologies, highlighting the urgent need for early and accurate ALL detection while exposing the limitations of conventional diagnostic approaches. This foundational analysis emphasized the necessity of leveraging deep learning, particularly convolutional neural networks and computer-aided diagnostic systems, to enhance diagnostic precision and efficiency.

The research proceeded to practical implementation, encompassing meticulous image preprocessing, the construction of sophisticated deep learning architectures, and rigorous performance evaluation. The results achieved, particularly the VGG19 model’s accuracy of 99.25\%, surpassed existing works on the same dataset, underscoring the effectiveness of the proposed approach and its potential to significantly improve automated ALL detection systems.

Moreover, this study underscores the critical importance of interdisciplinary collaboration between medical expertise and artificial intelligence. The synergistic integration of human clinical knowledge with cutting-edge machine learning algorithms holds the promise of enhancing diagnostic accuracy and efficiency, ultimately benefiting patient outcomes through more timely and reliable diagnostic procedures.

In essence, this research has contributed valuable insights and advancements in the field of ALL detection, paving the way for further exploration, refinement, and clinical integration of deep learning-based diagnostic systems. As technological innovation continues to evolve, the adoption of such systems in clinical practice holds immense potential to alleviate the diagnostic burden on healthcare professionals while enhancing patient care and survival rates.
\bibliographystyle{unsrt}

\bibliography{Reference}

\end{document}